\documentclass[10pt,twocolumn,letterpaper]{article}

\usepackage[pagenumbers]{utils/cvpr} 

\usepackage{siunitx}
\usepackage{booktabs} 

\usepackage{float} % For [H] float option

\usepackage{xspace}

\usepackage{makecell}
\usepackage{pifont}
\usepackage{multirow}
\definecolor{my_green}{RGB}{51,102,0}
\definecolor{my_red}{RGB}{204, 0, 0}
\definecolor{my_half}{RGB}{226,115,0} 
\renewcommand{\checkmark}{\textcolor{my_green}{\ding{51}}} % ✔
\newcommand{\crossmark}{\textcolor{my_red}{\ding{55}}} % ✘
\newcommand{\halfcheck}{\textcolor{my_half}{\ding{51}\rotatebox[origin=c]{-9.2}{\kern-0.7em\ding{55}}}}

\definecolor{duetT}{HTML}{5B4B8A}  
\definecolor{duetE}{HTML}{7B6BA8} 
\definecolor{duetX}{HTML}{9B8BC6}  
\definecolor{duetT}{HTML}{B8AAD9}  
\definecolor{duetD}{HTML}{E8956B}  
\definecolor{duetI}{HTML}{EA7A52} 
\definecolor{duetB}{HTML}{D65C3B} 
\definecolor{duetN}{HTML}{F65C3B}  
\definecolor{duetC}{HTML}{365C3B} 
\definecolor{duetH}{HTML}{665C3B}

\newcommand{\dataset}{%
  \textcolor{duetT}{T}%
  \textcolor{duetE}{e}%
  \textcolor{duetX}{x}%
  \textcolor{duetT}{t}%
  \textcolor{duetE}{E}%
  \textcolor{duetD}{d}%
  \textcolor{duetI}{i}%
  \textcolor{duetT}{t}%
  \textcolor{duetB}{B}%
  \textcolor{duetE}{e}%
  \textcolor{duetN}{n}%
  \textcolor{duetC}{c}%
  \textcolor{duetH}{h}%
  \xspace
}

\usepackage[titles]{tocloft}

\definecolor{cvprblue}{rgb}{0.21,0.49,0.74}
\usepackage[pagebackref,breaklinks,colorlinks,allcolors=cvprblue]{hyperref}

\def\paperID{} 
\def\confName{}
\def\confYear{}

\title{\textbf{\dataset: Evaluating Reasoning-aware Text Editing Beyond Rendering}}

\author{
    \textbf{Rui Gui}\textsuperscript{1,*} \quad
    \textbf{Yang Wan}\textsuperscript{1,*} \quad
    \textbf{Haochen Han}\textsuperscript{2,*} \quad
    \textbf{Dongxing Mao}\textsuperscript{1} \\[0.3em] 
    \textbf{Fangming Liu}\textsuperscript{2} \quad
    \textbf{Min Li}\textsuperscript{1} \quad
    \textbf{Alex Jinpeng Wang}\textsuperscript{1,$\dagger$} \\[0.8em] 
    \textsuperscript{1}Central South University \quad
    \textsuperscript{2}Pengcheng Laboratory \\[0.8em] 
    \small *Equal contribution. \quad $\dagger$Corresponding author.
}

\begin{document}

\twocolumn[{%
    \renewcommand\twocolumn[1][]{#1}%
    \maketitle
    \begin{center}
        \centering
        \captionsetup{type=figure} 
        \twocolumn
[{%
\renewcommand\twocolumn[1][]{#1}%
\maketitle
\begin{center}
    \centering
    \captionsetup{type=figure}
    \includegraphics[width=.9\linewidth]{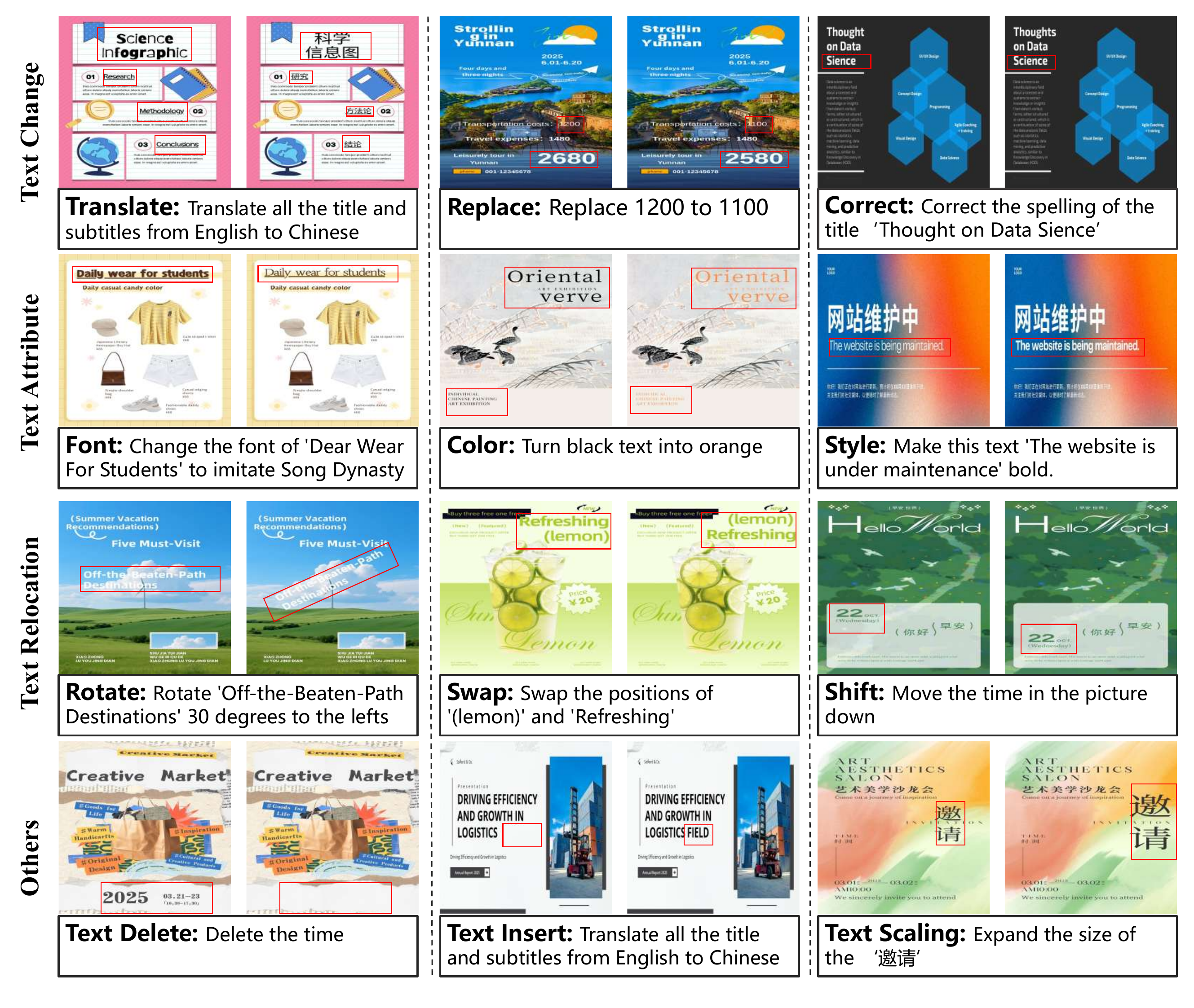}
    \captionof{figure}{
    \textbf{
    Overview of \dataset.}
    \dataset covers diverse text-in-image editing types such as translation, replacement, color and rotation adjustment, text removal, scaling, and creation, spanning both visual fidelity and \textit{reasoning-intensive semantic edits}.
    For clarity, we highlight the major regions of modification with red rectangle.
    }
    \label{fig:teaser}
\end{center}%
}]
 
    \end{center}%
}]

\clearpage 

\begin{abstract}
Text rendering has recently emerged as one of the most challenging frontiers in visual generation, drawing significant attention from large-scale diffusion and multimodal models.
However, text editing within images remains largely unexplored, as it requires generating legible characters while preserving semantic, geometric, and contextual coherence.
To fill this gap, we introduce \dataset, a comprehensive evaluation benchmark that explicitly focuses on text-centric regions in images.
Beyond basic pixel manipulations, our benchmark emphasizes reasoning-intensive editing scenarios that require models to understand physical plausibility, linguistic meaning, and cross-modal dependencies.
We further propose a novel evaluation dimension, \textbf{Semantic Expectation (SE)}, which measures reasoning ability of model to maintain semantic consistency, contextual coherence, and cross-modal alignment during text editing.
Extensive experiments on state-of-the-art editing systems reveal that while current models can follow simple textual instructions, they still struggle with context-dependent reasoning, physical consistency, and layout-aware integration. 
By focusing evaluation on this long-overlooked yet fundamental capability, \dataset establishes a new testing ground for advancing text-guided image editing and reasoning in multimodal generation.
\end{abstract}    
\section{Introduction}
\label{sec:intro_v2}

The ever-increasing volume of visual content has driven progress in text-guided image editing, which aims to modify images flexibly through natural language instructions. Recent advances in diffusion models \cite{kulikov2023sinddm} and large vision-language models \cite{wang2025multi} have even shown generation capabilities indistinguishable to human observers.

However, much of their success is confined to editing intuitive visual objects. When the target shifts to \textbf{text embedded in the image}, existing generative models exhibit notable quality degradation, as textual content possesses high semantic density and is tightly coupled with layout, typography, perspective, and scene context—severely hindering their applications in real-world scenarios, \eg, advertising customization and watermark manipulation.

Driven by the demand for precise text editing, text rendering\cite{glyph,glyph-v2,glyphcontrol,textdiffuser-2,glyphdraw,anytext,anytext2} has become a key metric for evaluating generative foundation models. For example, half of the GPT-4o image generation demonstrations focus on text rendering scenarios. To systematically assess this capability, a series of benchmarks \cite{lex-bench, shi2025wordcon,longtext-bench,textatlas5m,anytext} has been proposed to measure controllability at the visual text level. However, existing benchmarks primarily focus on basic pixel-level manipulations and overlook higher-level semantic challenges. Specifically, through empirical analysis of advanced models \cite{cui2025emu3,google2025nanobanana,gpt4o,qwenimagetechnicalreport,seedream2025seedream}, we observe three common failure modes: \textbf{(i) Text rendering and legibility.} Models may hallucinate characters, misspell words, or distort glyphs, deviating from the specified content.  
\textbf{(ii) Visual and stylistic consistency.} Inserted or replaced text often mismatches the background in font, color, perspective, or illumination, leading to “copy–paste’’ artifacts.  
\textbf{(iii) Semantic consistency.} Editing prices, labels, or chart values can break logical or factual relations in the scene, revealing limited reasoning ability.  
Therefore, there is an urgent need for a benchmark that requires models to reason over \emph{contextual and logical relations} among textual elements to preserve global coherence.

\begin{table*}[ht]
    \centering
    \normalsize
    \caption{\textbf{Comparison of Existing Datasets and \dataset.}}
    \label{tab:dataset_comparison}
    \resizebox{\textwidth}{!}{
        \begin{tabular}{l|ccc|cccc}
        \toprule 
        \textbf{Benchmark}  & \multicolumn{3}{c}{\bf Dataset Characteristics} & \multicolumn{4}{c}{\bf Model Capability Dimensions}\\
        \midrule
        & \textbf{Complexity}  
        & \textbf{\makecell[c]{Human Annotations}} 
        & \textbf{\makecell[c]{Multilingual}}
        & \textbf{\makecell[c]{Text Rendering}}
        & \textbf{\makecell[c]{Text Editing}}
        & \textbf{\makecell[c]{Layout Consistency}}
        & \textbf{\makecell[c]{Reasoning}} \\
        \midrule
        LAION-Glyph\cite{glyphcontrol} &easy&\crossmark&\crossmark&\checkmark&\crossmark&\crossmark&\crossmark  \\
        \midrule
        AnyText\cite{anytext} &easy&\crossmark&\checkmark&\checkmark&\crossmark&\crossmark&\crossmark  \\
        \midrule
        LeX-Bench\cite{lex-bench} &diverse&\crossmark&\crossmark&\crossmark&\crossmark&\checkmark&\crossmark  \\
        \midrule
        \dataset&diverse&\checkmark&\checkmark&\checkmark&\checkmark&\checkmark&\checkmark  \\
        \bottomrule
        \end{tabular}
    }
\end{table*}

In this work, we propose \dataset to unify the evaluation of editing textual content within images, \textbf{textual accuracy}, \textbf{visual realism}, and \textbf{reasoning consistency} for text-in-image editing. 
As shown in Table~\ref{tab:dataset_comparison}, our benchmark uniquely covers all key dimensions of text-in-image editing, including multilinguality, human annotations, layout consistency, and reasoning.
\dataset covers 14 topics, 6 task types, and 12 fine-grained sub-tasks, with 1196 annotated instances featuring complex layouts, multilingual content, and challenging surfaces.  

Our evaluation methodology follows two complementary tracks:  
(1) MLLM acts as an evaluator to score five dimensions—Instruction Following (IF), Text Accuracy (TA), Visual Consistency (VC), Layout Preservation (LP), and Semantic Expectation (SE); SE can optionally leverage a \emph{knowledge prompt} (\emph{kp}) to make implicit semantic dependencies explicit. 
(2) A reproducible script computes perceptual metrics—\textbf{SSIM}, \textbf{PSNR}, \textbf{MSE}, and masked-region evaluation metric \textbf{LPIPS}.  
The two tracks are reported side by side and can be aggregated into an overall score.

We list our contribution as below:
(1) We introduce \dataset, the \textbf{first benchmark tailored to text-in-image} editing, spanning 14 topics, 6 task types, 12 sub-tasks, and 1{,}400 annotated instances.   
(2) We introduce the \textbf{Semantic Expectation (SE)} metric to probe knowledge-supported and cross-element dependencies in textual edits; results on recent systems reveal persistent gaps in controllability, visual integration, and reasoning alignment.

\section{Related Work}
\label{sec:related_work}

\paragraph{Text-Conditioned Image Editing.}
Recent years have witnessed rapid progress in text-guided image editing, driven by the evolution of diffusion models and multimodal large language models (MLLMs).  
Early modular systems such as InstructPix2Pix~\cite{brooks2023instructpix2pix}, InstructEdit~\cite{wang2023instructeditimprovingautomaticmasks}, and MagicBrush~\cite{zhang2024magicbrushmanuallyannotateddataset} interpret textual instructions to produce intermediate masks or latent representations for diffusion-based editing.  
Subsequent works including AnyEdit~\cite{yu2025anyedit}, UltraEdit~\cite{zhao2024ultraeditinstructionbasedfinegrainedimage},Wordcon\cite{shi2025wordcon} and SmartEdit~\cite{huang2023smarteditexploringcomplexinstructionbased} further improve instruction following and fine-grained control.  
Meanwhile, unified multimodal frameworks such as OmniGen~\cite{xiao2024omnigenunifiedimagegeneration}, ACE/ACE++~\cite{han2024aceallroundcreatoreditor,mao2025aceinstructionbasedimagecreation}, bagel\cite{deng2025emerging}, Emu3.5\cite{cui2025emu3}demonstrate the growing capability of generalist systems to perform editing as part of broader generative reasoning.
Despite these advances, most existing approaches still treat text within images as generic texture rather than structured symbols, leading to spelling errors, poor stylistic integration, and background artifacts—revealing a lack of semantic and glyph-level control.

\paragraph{Datasets for Image Editing.}
Progress in instruction-based editing has been largely enabled by large-scale datasets such as InstructPix2Pix~\cite{brooks2023instructpix2pix}
, MagicBrush~\cite{zhang2024magicbrushmanuallyannotateddataset}, HQ-Edit~\cite{hui2024hqedithighqualitydatasetinstructionbased}, Infographics-650K\cite{peng2025bizgen}and SEED-Data-Edit~\cite{ge2024seeddataedittechnicalreporthybrid}.  
However, these resources focus primarily on object- or style-level edits, lacking the fine-grained alignment and linguistic structure needed for text-in-image manipulation.  
Datasets from text recognition or captioning (e.g., COCO-Text~\cite{veit2016cocotextdatasetbenchmarktext}, TextCaps~\cite{sidorov2020textcapsdatasetimagecaptioning}) provide text understanding but not paired (instruction, target) samples for editing.  
This motivates the need for a dedicated benchmark addressing both perceptual fidelity and semantic reasoning in textual edits.

\paragraph{Benchmarks and Evaluation Protocols.}
In recent years, several benchmark\cite{rise,wise,wang2025genexam,han2025unireditbench,wang2025unigenbench++,designt2i,cheng2025comt} tests have explored methods for evaluating reasoning in image generation and editing. These tests focus on types of reasoning like factual, causal, and logical reasoning. For instance, WISE\cite{wise} and RISEBench \cite{rise}assess these reasoning types in visual generation, while Complex-Edit \cite{yang2025complexedit}introduces adjustable editing chains to test reasoning in incremental image modifications. However, these studies often overlook the unique challenges of text editing within images.
In contrast, Lex-Bench \cite{lex-bench}evaluates text rendering performance in text-to-image generation with a new metric, PNED (Text Rendering Performance). TextAtlasEval\cite{textatlas5m} focuses on the rendering of long-form text, examining how models handle longer text. Yet, these methods fail to address the most challenging task in image editing: dynamic text modifications in complex contexts.
To fill this gap, TE-Bench was developed as a unified, user-centered framework for evaluating text editing in images. It combines perceptual and linguistic metrics and introduces a new reasoning dimension—Semantic Expectation (SE)—to assess models' understanding of semantic, stylistic, and contextual relationships. This approach bridges the gap between visual fidelity and multimodal reasoning, enabling more precise evaluation of models' abilities in complex text editing tasks.

\begin{figure}[t]
  \centering
  \includegraphics[width=.8\linewidth]{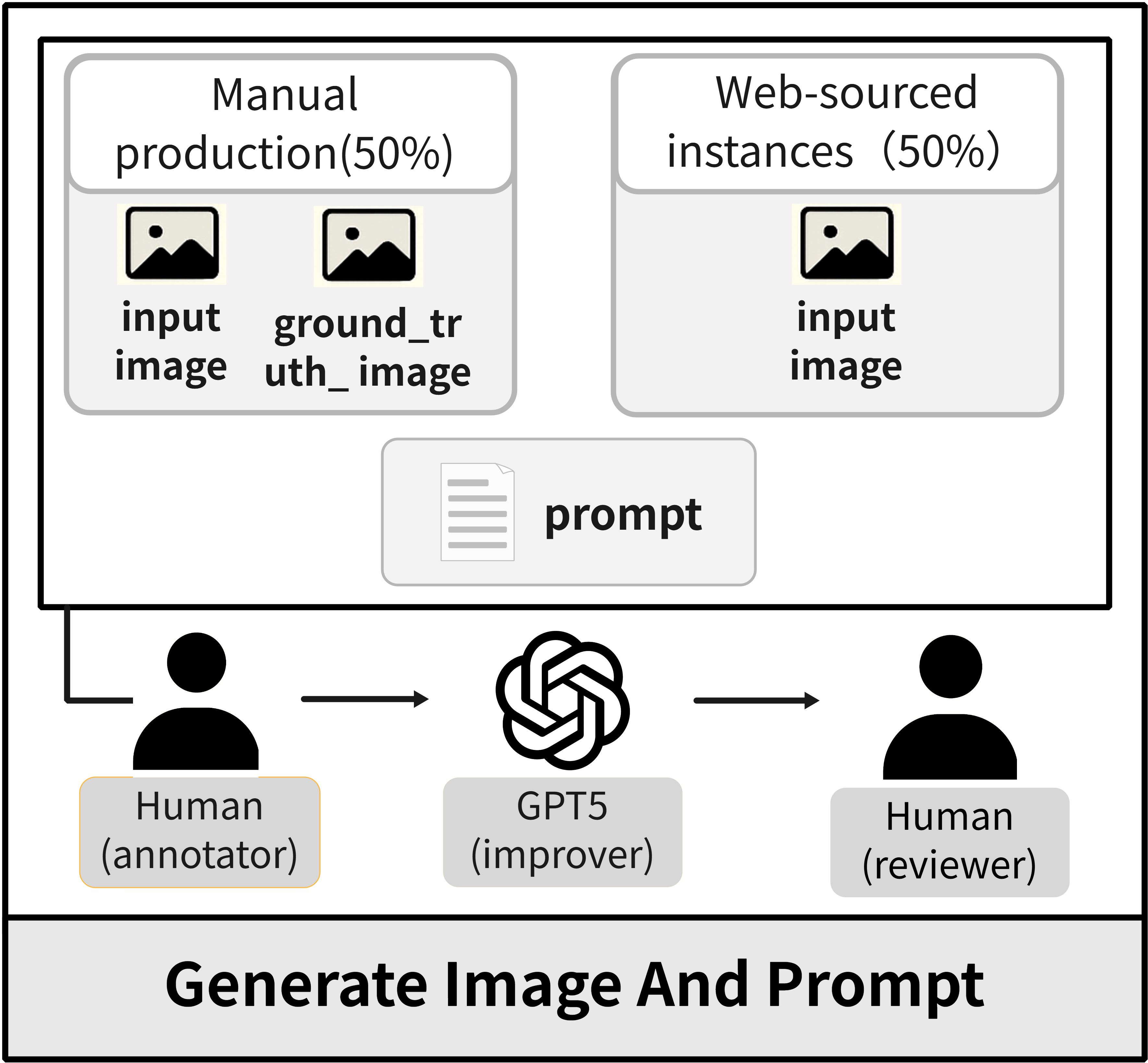}
  \caption{
  \textbf{Data collection and annotation pipeline of \dataset.}
  The dataset is constructed from a balanced mixture of manually produced instances with paired input/ground-truth images and web-sourced instances with input images only.
  For each sample, a human annotator first designs the editing instruction and attribute labels; a customized GPT5 prompt then refines and normalizes the text; finally, a human reviewer validates the outputs before inclusion in the benchmark.
  }
  \label{fig:pipeline}
\end{figure}

\section{Text Editing Benchmark}
\label{sec:benchmark}
Beyond surface-level rendering, our \dataset explicitly requires models to reason over semantic and contextual constraints to produce coherent edits. 
The overall benchmark construction pipeline is shown in Figure~\ref{fig:pipeline}.

\subsection{Data distribution}
\begin{figure}[t]
  \centering
  \includegraphics[width=.8\linewidth]{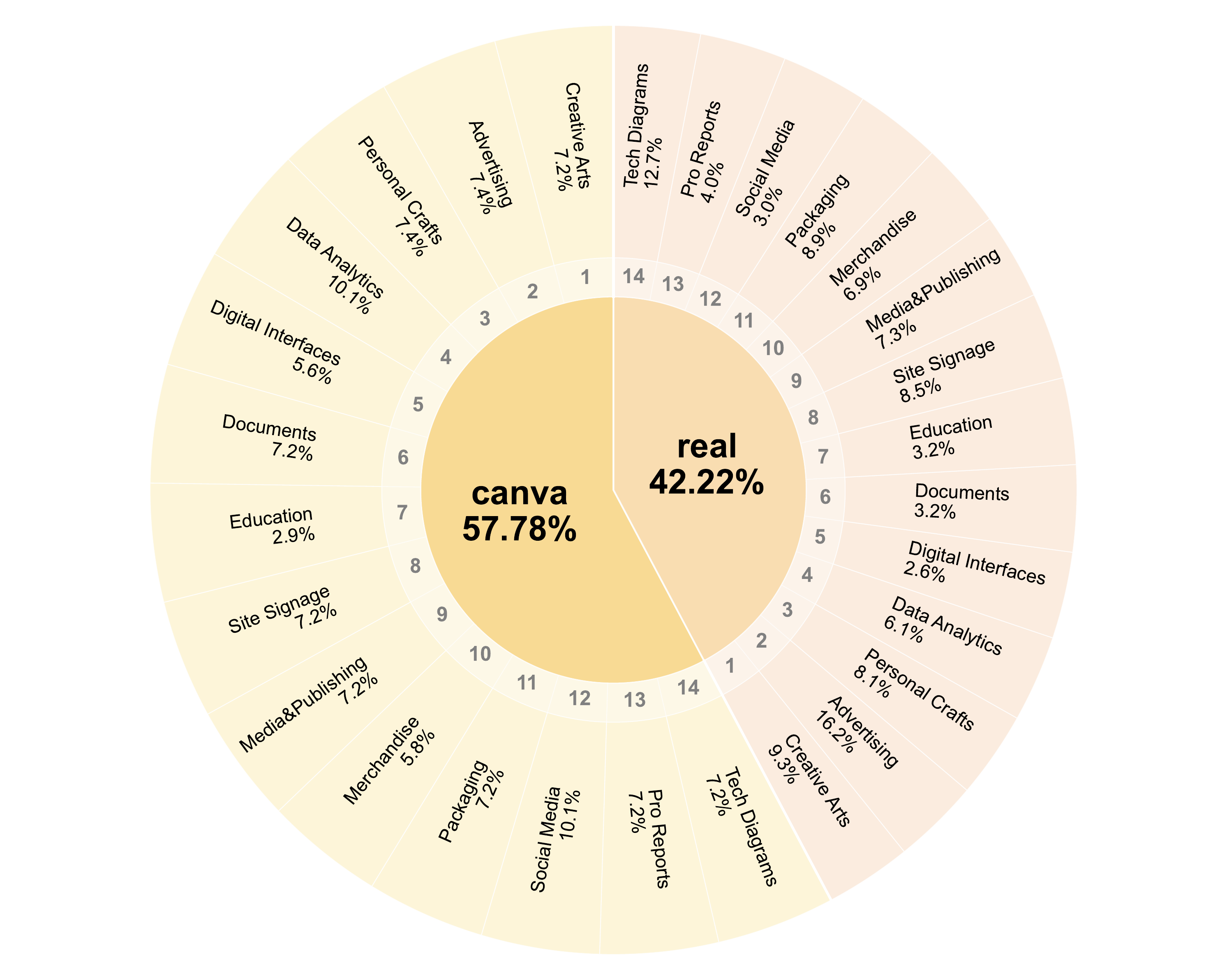}
  \caption{
    \textbf{Data distribution.}
    \dataset includes synthetic and real-world images, covering common text-in-image cases.
  }
  \label{fig:Data distribution}
\end{figure}

\dataset covers a broad set of real-world application domains, incorporating both human-synthesized data created via Canva~\footnote{https://www.canva.com} and human-annotated data sourced from real-world images. To ensure diversity, we engaged domain experts to curate 14 subject topics that frequently involve textual content in daily visual contexts (e.g., signage and documents). This yields 14 subject types for each split. The overall data composition is illustrated in Figure~\ref{fig:Data distribution}, demonstrating balanced coverage across these domains.

\subsection{Atom Operations}
\label{subsec:subjects}
For systematic evaluation, we define \textbf{a set of canonical editing operations} that cover common text-related edits encountered in these domains:
\textbf{Text Delete}: remove specified textual elements, emphasizing background recovery and contextual coherence.
\textbf{Text Insert}: introduces new textual elements into an existing layout, reflecting compositional and spatial reasoning capabilities.
\textbf{Text Change}: replace existing text while maintaining stylistic and spatial consistency.
\textbf{Text Relocation}: repositioning textual elements across spatial coordinates, including translations and rotations within the layout.
\textbf{Scaling}: adjusts the size of textual instances while preserving geometric and typographic balance.
\textbf{Text Attribute}: alters visual or stylistic characteristics such as font, weight, or color to achieve target appearance transformations.

\subsection{Difficulty Annotation}
\label{sec:difficulty}

To enable fine-grained analysis and controlled sampling, each instance is annotated with 
\emph{i.} a categorical difficulty label (\texttt{easy}/\texttt{medium}/\texttt{hard}) 
and \emph{ii.} a continuous difficulty score $\in [0,20]$.
The continuous score is computed as the sum of \textbf{ten interpretable difficulty attributes}, 
each rated in $\{0,1,2\}$.
Operational definitions of all attributes are summarized in Table~\ref{tab:difficulty_attributes}, 
and detailed annotation criteria are provided in the supplementary material.
The summed \texttt{difficulty\_score} ranges from 0 to 20, 
which is further mapped to discrete tiers using conservative thresholds:
\texttt{easy} (0–5), \texttt{medium} (6–10), and \texttt{hard} (11–20).

\begin{table}[!t]
\centering
\caption{Summary of difficulty attributes used for instance-level annotation. 
Each attribute is scored in $\{0,1,2\}$, contributing to the overall \texttt{difficulty\_score} $\in [0,20]$.}
\vspace{2mm}
\resizebox{\linewidth}{!}{
\begin{tabular}{ll}
\toprule
\textbf{Attribute} & \textbf{Description (score basis)} \\ 
\midrule
\texttt{num\_text\_regions} & Number of distinct text regions (1=0; 2–3=1; $>$3=2) \\
\texttt{text\_length} & Target text length (short$\le$2w=0; medium=1; long$>$4w=2) \\
\texttt{font\_complexity} & Typographic complexity (plain=0; serif=1; decorative=2) \\
\texttt{language} & Linguistic complexity (English=0; non-English=1; mixed=2) \\
\texttt{surface\_geometry} & Surface curvature (planar=0; mild=1; strong=2) \\
\texttt{occlusion} & Degree of occlusion (none=0; partial=1; severe=2) \\
\texttt{context\_dependency} & Global context required (no=0; yes=2) \\
\texttt{background\_clutter} & Background complexity (clean=0; moderate=1; cluttered=2) \\
\texttt{task\_category} & Operation type (simple=0; geometric=1; semantic=2) \\
\texttt{semantic\_linkage} & Cross-element reasoning required (no=0; yes=2) \\
\bottomrule
\end{tabular}}
\label{tab:difficulty_attributes}
\end{table}

\subsection{Data collection}
\label{subsec:data}
The overall data generation pipeline is shown in Figure~\ref{fig:pipeline}.
\dataset comprises 1,196 annotated instances. 
Data were collected and prepared through a two-pronged strategy to balance diversity and annotation fidelity:

\textbf{i. Manual production (57.78\%).} Half of the dataset was created by skilled annotators using commercial design tools. These instances include both the input image and an output image representing an ideal, manually produced edit. Providing paired ground-truth outputs facilitates supervised evaluation and pixel-level comparisons for this subset.

\textbf{ii. Web-sourced instances (42.22\%).}
To ensure sufficient scale and content diversity while maintaining annotation quality, the remaining half of the dataset was primarily derived from two public benchmarks, 
\textbf{GIEBench}\cite{wang2024giebench} and\textbf{AnyEdit}\cite{yu2025anyedit}.  
We selectively extracted images that met our filtering criteria (e.g., containing clearly localizable editable text regions and adequate contextual information).  
For categories or scenarios underrepresented in these sources, annotators were instructed to collect additional images from public image repositories and design-sharing websites.  
All selected images were subsequently normalized and annotated following the same pipeline and criteria as the manually produced subset.

To ensure annotation quality and consistency, we employ a structured annotation framework detailed below:

\noindent\textbf{Annotation Format.} 
Each instance comprises four components: \emph{i.} basic attribute descriptions, \emph{ii.} the human-written editing instruction, \emph{iii.} meta attributes for difficulty scoring, and \emph{iv.} a knowledge prompt (kp) when semantic linkage is present. 
The kp articulates the implicit semantic expectation and the reasoning chain that links the requested textual edit to the expected visual or contextual outcome.

\noindent\textbf{Annotation pipeline.} 
All annotations in \dataset are produced through a \textbf{three-stage human collaborative workflow}. \emph{i.} \textbf{Human Drafting:} Annotators produce the initial prompts, difficulty attributes, and kp drafts. 
\emph{ii.} \textbf{Model Refinement:} A customized GPT5~\cite{hurst2024gpt} prompt is then applied to normalize the text, unify the phrasing style, and produce a consistent kp structure. 
\emph{iii.} \textbf{Human Verification} senior annotators review and correct every item; only after this manual verification is a sample included in \dataset.

\section{Dual-Track Evaluation Framework}
\label{sec:metrics}

\begin{figure}[t]
    \centering
    \includegraphics[width=\linewidth]{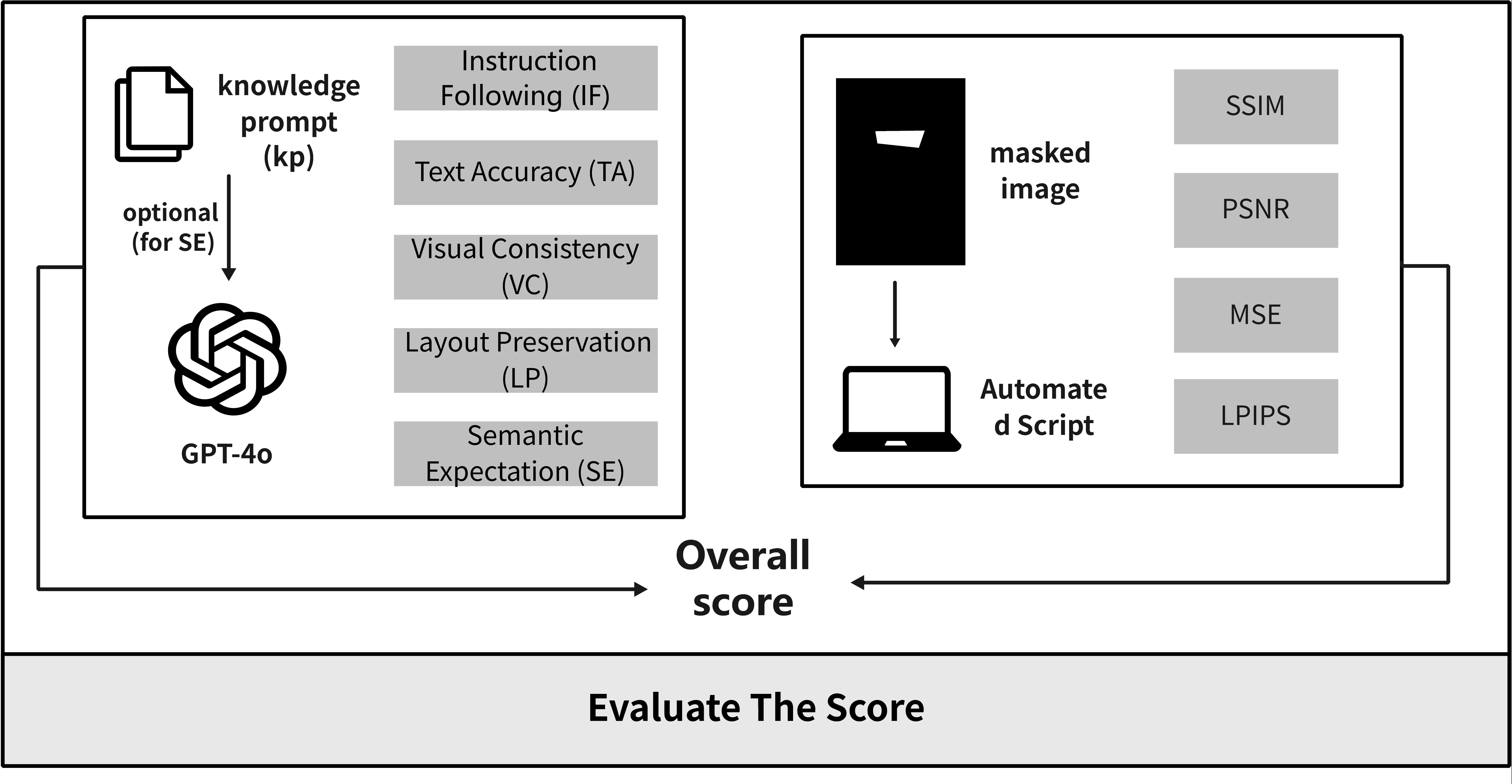}
    \caption{
    \textbf{Overview of the evaluation pipeline.}
    The evaluation framework combines both \textbf{Semantic Metrics} and \textbf{Pixel-Level Fidelity Metrics}.
    Left: GPT-4o  evaluates text-guided edits across five complementary dimensions.
    When the edit involves higher-level semantic reasoning, an auxiliary knowledge prompt is optionally provided for SE assessment.
    Right: Pixel-level objective metrics are proposed to quantify preservation of unedited content.
    }
    \label{fig:evaluation_pipeline}
\end{figure}

Evaluating text-in-image editing requires assessing both low-level visual fidelity and high-level semantic coherence. To address this, \dataset employs a \textit{dual-track evaluation framework} (Figure~\ref{fig:evaluation_pipeline}): \textbf{Pixel-Level Evaluation} uses human-annotated masks to isolate edited regions and measure preservation fidelity objectively, while \textbf{Semantic Evaluation} leverages multimodal LLMs to verify reasoning consistency and contextual understanding.

\subsection{ Pixel-Level Objective Metrics}
To evaluate the visual fidelity of edited images, we employ a set of \textbf{pixel-level objective metrics} that quantify how well models \textit{preserve unedited background regions}. 
Since existing automatic segmentation methods are often inaccurate for text-centric edits, all editable regions in \dataset are \textbf{manually annotated} by trained annotators to ensure pixel-level precision.

\noindent\textbf{Human-Annotated Masks.}  
For each instance, we define an editable mask that precisely delineates the region intended for modification. The annotation strategy adapts to data availability: when ground-truth edited images exist (e.g., the Canva subset), masks are aligned with regions that differ between the input and reference images. For real-world samples lacking paired outputs, annotators manually delineate target regions based on textual instructions, using the original image as a reference. This human-in-the-loop approach ensures reliable separation of edited and static regions, enabling reproducible quantitative evaluation.

\noindent\textbf{Evaluation Protocol.}  
To measure background preservation fidelity, metrics are computed exclusively on \textbf{non-edited regions}. For samples with ground-truth edits (Canva subset), we compare model outputs against reference images, excluding masked regions. For real-world samples without paired outputs, we compare generated images against original inputs outside masked areas, quantifying the extent of unintended modifications to preserved content.

\noindent \textbf{Metric Implementation Details.}
Using the annotated masks and evaluation protocol, we measure background preservation with multiple metrics.
Our primary metric is a calibrated masked MSE that measures unintended alterations in non-
edited regions. 
To account for spatial misalignment introduced during editing, we apply SIFT keypoint detection\cite{lowe2004distinctive}, FLANN-based matching\cite{muja2009fast}, and affine transformation for robust alignment. 
We additionally compute masked SSIM, LPIPS, and PSNR to jointly assess perceptual and pixel-level preservation quality.

\subsection{ MLLM-based Semantic  Metrics}
While pixel-level metrics assess local fidelity, they cannot capture whether edits fulfill semantic intent or maintain global coherence. To address this gap, we employ multimodal large language models (MLLMs) to evaluate five complementary semantic dimensions:
\textit{Instruction Following, Text Accuracy, Visual Consistency, Layout Preservation, and Semantic Expectation}.
These metrics jointly assess both \textbf{low-level editing fidelity} and \textbf{high-level semantic alignment}, providing a holistic evaluation of reasoning-aware editing capabilities.
Each metric is rated on a 0–5 scale using GPT-4o~\cite{gpt4o} as the evaluator.

\subsubsection{Overview of Semantic Evaluation}
Specifically, \textbf{Instruction Following (IF)} checks whether the model executes the instruction exactly as described, without adding or altering anything unintended.
\textbf{Text Accuracy (TA)} measures the correctness and completeness of textual modifications, ensuring that the edited text faithfully reflects the intended meaning of the instruction.
\textbf{Visual Consistency (VC)} assesses whether the newly edited content blends seamlessly into the surrounding visual context, including font style, color tone, and illumination, thereby reflecting perceptual coherence and realism.
\textbf{Layout Preservation (LP)} examines whether edits remain spatially localized and whether non-target regions preserve their structural integrity, avoiding unnecessary distortions that compromise the overall layout.
Finally, we introduce \textbf{Semantic Expectation (SE)} as a novel metric for evaluating higher-order reasoning and contextual understanding beyond surface-level text modification.

\begin{figure*}[!htbp]
    \centering
    \includegraphics[width=\linewidth]{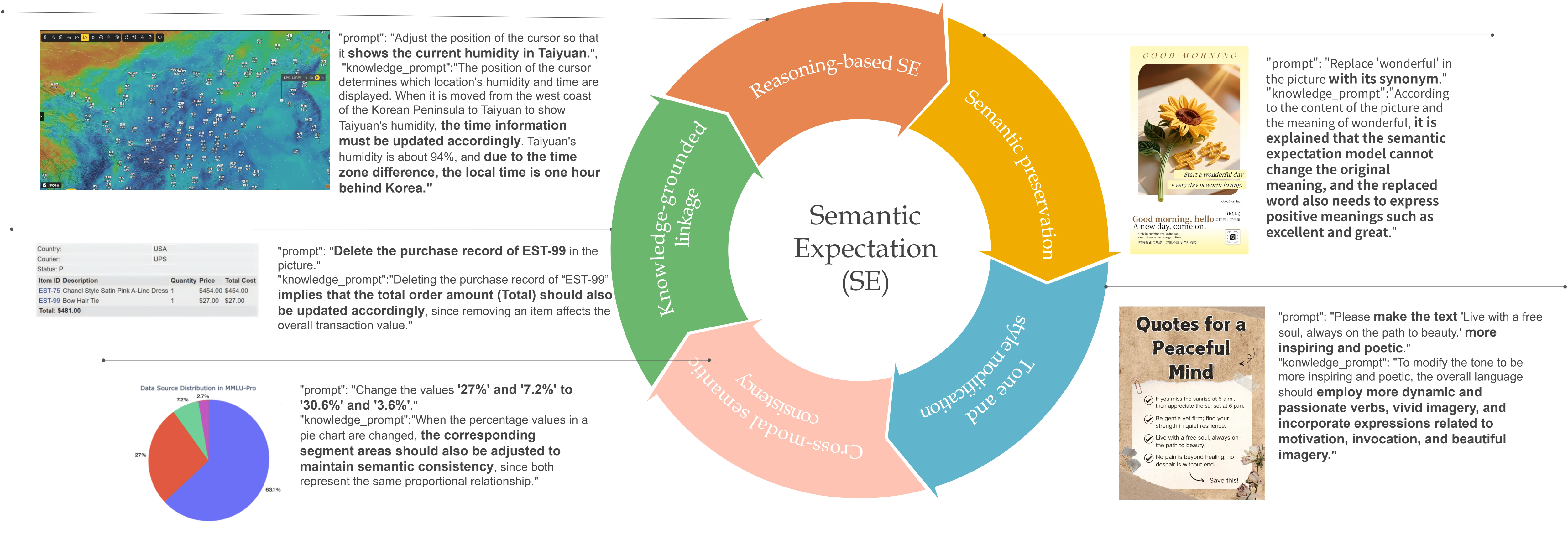}
    \caption{
    Illustration of the five dimensions of the \textbf{Semantic Expectation (SE)} metric. 
    Key conceptual aspects are highlighted for clarity.}
    \label{fig:se_metric}
\end{figure*}

\subsubsection{Semantic Expectation (SE) Metric}
\label{sec:metrics-se}

While the previous four metrics assess local fidelity and explicit textual alignment,
\textbf{Semantic Expectation (SE)} evaluates models' deeper reasoning and contextual understanding capabilities.
SE measures whether models can infer and apply \emph{implicit semantic dependencies} between textual instructions and corresponding visual or contextual outcomes.
In Figure~\ref{fig:se_metric},we show complementary dimensions of SE:

\emph{i.} \textbf{Knowledge-grounded linkage:} Evaluates whether the model can leverage commonsense or domain-specific knowledge implied by the instruction.
\emph{ii.} \textbf{Reasoning-based SE:} Measures the ability to perform implicit logical or causal reasoning across dependent elements, ensuring internal semantic consistency.
\emph{iii.} \textbf{Semantic preservation:} Checks whether the model can modify text while preserving its meaning.
\emph{iv.} \textbf{Tone and style modification:} Evaluates the ability to adjust tone or communicative intent without distorting factual semantics.
\emph{v.} \textbf{Cross-modal semantic consistency:} Examines whether textual edits remain coherent with surrounding visual cues or environmental context, ensuring cross-modal alignment.

To evaluate \textbf{Semantic Expectation (SE)},
we follow \textbf{KRIS-Bench}~\cite{wu2025kris} and introduce an auxiliary component termed \textbf{Knowledge Prompt (KP)}.
Each \emph{KP} explicitly articulates the implicit semantic expectation within the instruction and the reasoning chain linking the textual edit to its expected visual or contextual outcome.
By providing explicit human-interpretable rationale, the \emph{KP} improves GPT-4o’s accuracy and consistency in assessing reasoning.

\begin{table*}[t] 
\centering
\caption{
\textbf{Comprehensive Evaluation Results on TE-Bench (Part 1: Script-based).}
Each model is evaluated on both \textit{Synthetic} and \textit{Real-World} subsets.
This table shows the automated \textbf{Script-based Evaluation (S)}.
Higher is better for all metrics except MSE.
}
\renewcommand{\arraystretch}{0.9} 
\resizebox{.95\textwidth}{!}{ 
\begin{tabular}{l | rrrr | rrrr} 
\toprule
\multirow{3}{*}{\textbf{Model}} & \multicolumn{8}{c}{\textbf{Script-based Evaluation (S)}} \\
\cmidrule(lr){2-9}
& \multicolumn{4}{c|}{\textbf{Synthetic}} & \multicolumn{4}{c}{\textbf{Real-World}} \\
\cmidrule(lr){2-5} \cmidrule(lr){6-9}
& \textbf{SSIM$\uparrow$} & \textbf{LPIPS$\downarrow$} & \textbf{PSNR$\uparrow$} & \textbf{MSE$\downarrow$} 
& \textbf{SSIM$\uparrow$} & \textbf{LPIPS$\downarrow$} & \textbf{PSNR$\uparrow$} & \textbf{MSE$\downarrow$} \\
\midrule
Step1X-Edit & 0.879 & 0.089 & 25.523 & 982.548 & 0.872 & 0.091 & 27.169 & 997.937 \\
Step1X-Edit-Think & 0.899 & 0.067 & 27.012 & 584.159 & 0.861 & 0.099 & 26.710 & 1319.081 \\
MagicBrush & 0.788 & 0.162 & 19.292 & 3456.709 & 0.746 & 0.186 & 19.722 & 3646.180 \\
InstructPix2Pix & 0.768 & 0.187 & 17.774 & 2718.524 & 0.725 & 0.245 & 16.937 & 3601.554 \\
Emu3.5(512) & 0.769 & 0.092 & 19.323 & 1131.326 & 0.758 & 0.082 & 21.633 & 842.310 \\
OmniGen2 & 0.836 & 0.129 & 21.376 & 2046.597 & 0.818 & 0.140 & 22.726 & 2039.790 \\
Bagel(512) & 0.873 & 0.077 & 24.423 & 726.406 & 0.894 & 0.063 & 28.187 & 545.238 \\
Bagel-Think(512) & 0.896 & 0.057 & 25.524 & 459.235 & \textbf{0.901} & 0.058 & 28.364 & 423.021 \\
FLUX.1-Kontext-dev & \textbf{0.906} & 0.056 & \underline{28.290} & 522.070 & \underline{0.896} & 0.069 & 29.637 & 399.199 \\
Qwen-Image-Edit & 0.901 & \underline{0.039} & 28.219 & \underline{278.521} & 0.887 & \underline{0.045} & \underline{30.023} & \underline{228.439} \\
NanoBanana & \underline{0.904} & \textbf{0.036} & \textbf{29.883} &\textbf{160.341}  & 0.888 &\textbf{ 0.043} & \textbf{31.476} & \textbf{175.517} \\
Seedream & 0.775 & 0.136 & 20.329 & 1121.807 & 0.808 & 0.143 & 24.229 & 496.756\\
\midrule
Mean & 0.850 & 0.094 & 23.914 & 1182.354 & 0.838 & 0.105 & 25.568 & 1226.252  \\
\bottomrule
\end{tabular}
} 
\label{tab:tebench_script_eval_wide} 
\end{table*}

\newcolumntype{d}{S[table-format=2.2]}
\begin{table*}[t] 
\centering
\caption{
\textbf{Comprehensive Evaluation Results on TE-Bench (Part 2: MLLM-Assisted).}
This table shows the \textbf{GPT-4o–Assisted Evaluation (G) over 5 metrics}.
Each metric is in range 0 to 5.
}
\renewcommand{\arraystretch}{0.9} 

\newcolumntype{d}{S[table-format=2.2]}
\resizebox{.95\textwidth}{!}{ 
\begin{tabular}{l | rrrrrr | rrrrrr} 
\toprule
\multirow{3}{*}{\textbf{Model}} & \multicolumn{12}{c}{\textbf{GPT-4o–Assisted Evaluation (G)}} \\
\cmidrule(lr){2-13}
& \multicolumn{6}{c|}{\textbf{Synthetic}} & \multicolumn{6}{c}{\textbf{Real-World}} \\
\cmidrule(lr){2-7} \cmidrule(lr){8-13}

& \textbf{IF$\uparrow$} & \textbf{TA$\uparrow$} & \textbf{VC$\uparrow$} & \textbf{LP$\uparrow$}& \textbf{SE$\uparrow$} &\textbf{Overall$\uparrow$}
& \textbf{IF$\uparrow$} & \textbf{TA$\uparrow$} & \textbf{VC$\uparrow$} & \textbf{LP$\uparrow$}& \textbf{SE$\uparrow$} &\textbf{Overall$\uparrow$} \\
\midrule
Step1X-Edit & 1.40 & 1.44 & 1.88 & 3.40& 1.14 & 9.26/25& 1.96 & 2.11 & 2.81 & 4.06 & 1.08 & 12.02/25 \\
Step1X-Edit-Think & 2.28 & 2.50 & 2.52 & 3.81 & 2.07 & 13.18/25& 3.05 & 3.42 & 3.48 & 4.24 &1.98  & 16.17/25\\
MagicBrush & 0.73 & 0.73 & 0.60 & 1.18 & 0.62&3.86/25 & 0.59 & 0.80 & 0.64 & 1.42 & 0.67  & 4.12/25\\
InstructPix2Pix & 0.90 & 1.14 & 1.14 & 1.61 & 0.72 &5.51/25& 0.56 & 1.22 & 1.09 & 1.69 & 0.79 & 5.35/25 \\
Emu3.5(512) & 1.40 & 2.41 & 1.98 & 1.90 & 1.65 &9.34/25& 2.82 & 3.00 & 3.10 & 3.73 & 1.40  & 14.05/25\\
OmniGen2 & 1.12 & 1.74 & 1.72 & 3.08 & 0.86 &8.52/25& 1.67 & 2.22 & 2.45 & 3.63 & 1.02 & 10.99/25 \\
Bagel(512) & 1.81 & 2.31 & 1.83 & 2.94 & 1.52 &10.41/25& 2.21 & 2.50 & 2.75 & 4.22 & 1.15 & 12.83/25 \\  
Bagel-Think(512) & 1.53 & 1.85 & 1.67 & 2.97 & 1.34 &9.36/25& 1.53 & 1.53 & 2.13 & 4.09 & 0.93  & 10.21/25\\
FLUX.1-Kontext-dev & 1.85 & 2.15 & 2.67 & 4.35 & 1.40 &12.42/25 & 2.53 & 2.94 & 3.57 & 4.66 & 1.23 & 14.93/25\\
Qwen-Image-Edit & \textbf{2.91} & 3.23 & \textbf{3.51} & \textbf{4.48} & 2.45 & \textbf{16.58/25}& \underline{3.50} & \underline{3.83} & \textbf{4.15} & \underline{4.75} & 2.47 & \textbf{18.70/25} \\
NanoBanana & \underline{2.90} & \underline{3.25} & \underline{3.40} & \underline{4.46} & \underline{2.53} & \underline{16.54/25}& 3.18 & 3.60 & \underline{3.94} & \textbf{4.77} & \textbf{2.73}  & 18.22/25\\
Seedream & 2.67 & \textbf{3.44} & 2.97 & 3.25 & \textbf{2.57} & 14.90/25 &\textbf{3.64} & \textbf{3.96} & 3.90 & 4.42 & \underline{2.62} & \underline{18.54/25}\\
\midrule
Mean & 1.79 & 2.18 & 2.16 & 3.12 & 1.57 & 10.82/25 & 2.27 & 2.59 & 2.83 & 3.81 & 1.51 & 13.01/25  \\
\bottomrule
\end{tabular}
} 
\label{tab:tebench_gpt_eval_wide} 
\end{table*}

\section{Experiments}
\label{sec:experiments}

We evaluate a diverse set of text-guided image editing models on the \dataset to provide a comprehensive comparison between open-source and proprietary systems.
All experiments are conducted on eight NVIDIA Tesla V100 GPUs using  official default hyperparameter configuration of each model to ensure fairness and reproducibility.

\subsection{Baselines}
\textbf{Models.} We consider \textbf{eleven} representative text-guided image editing models covering a wide range of architectures and training paradigms.
\textbf{Open-source models:} Step1X-Edit~\cite{liu2025step1x}, MagicBrush~\cite{zhang2023magicbrush}, InstructPix2Pix~\cite{brooks2023instructpix2pix}, OmniGen2~\cite{wu2025omnigen2}, Emu3.5~\cite{cui2025emu3}, Bagel~\cite{deng2025emerging}, FLUX.1-Kontext-dev~\cite{labs2025flux}, and Qwen-Image-Edit~\cite{wu2025qwen}.
\textbf{Proprietary (closed-source) models:} NanoBanana~\cite{google2025nanobanana}, and Seedream~\cite{seedream2025seedream}.
By defaults, for each model family, we use the largest and best-performing available model to data.
These models collectively represent the state of the art in diffusion-based, multimodal large–model–based, and instruction-tuned architectures for image editing.

\noindent \textbf{Evaluation.} The result of each model is assessed through a dual evaluation scheme combining \textbf{Pixel-Level Objective Metrics} and \textbf{MLLM-based Semantic Metrics}:
This hybrid evaluation strategy balances objective reproducibility and subjective semantic fidelity, providing a rigorous, multi-perspective assessment of text editing capability.

\subsection{Quantitative Results}
\label{subsec:results}

This section presents a unified quantitative comparison across all models and both evaluation sets.
We shown the results in Table~\ref{tab:tebench_script_eval_wide} that derived from automated \textbf{pixel-level objective evaluation}, while and Table~\ref{tab:tebench_gpt_eval_wide} are obtained from \textbf{MLLM–assisted human-aligned scoring}.
Each model is evaluated on both the \textit{synthetic} and \textit{Real}  subsets.

We find:
\emph{i}. The benchmark is quite challenge for existing models no matter open-source or closed-source models.
For example, for best NanoBanana model, it only get 16.54 score in total, far away from 25.
\emph{ii}. The \textit{Synthetic} split is much harder than the \textit{Real-World} split, as evidenced by the mean value (10.82 vs 13.01).
\emph{iii}. The best metric is LP, which get 3.12/5 score average, this indicates most models can preserve the layout well during text editing process.
\emph{iv}. The worst metric is Semantic Expectation, indicating that most models fail to perform the required reasoning. 
This is expected, as our benchmark requires multi-step reasoning grounded in world knowledge.

\begin{figure*}
    \centering
\includegraphics[width=\linewidth]{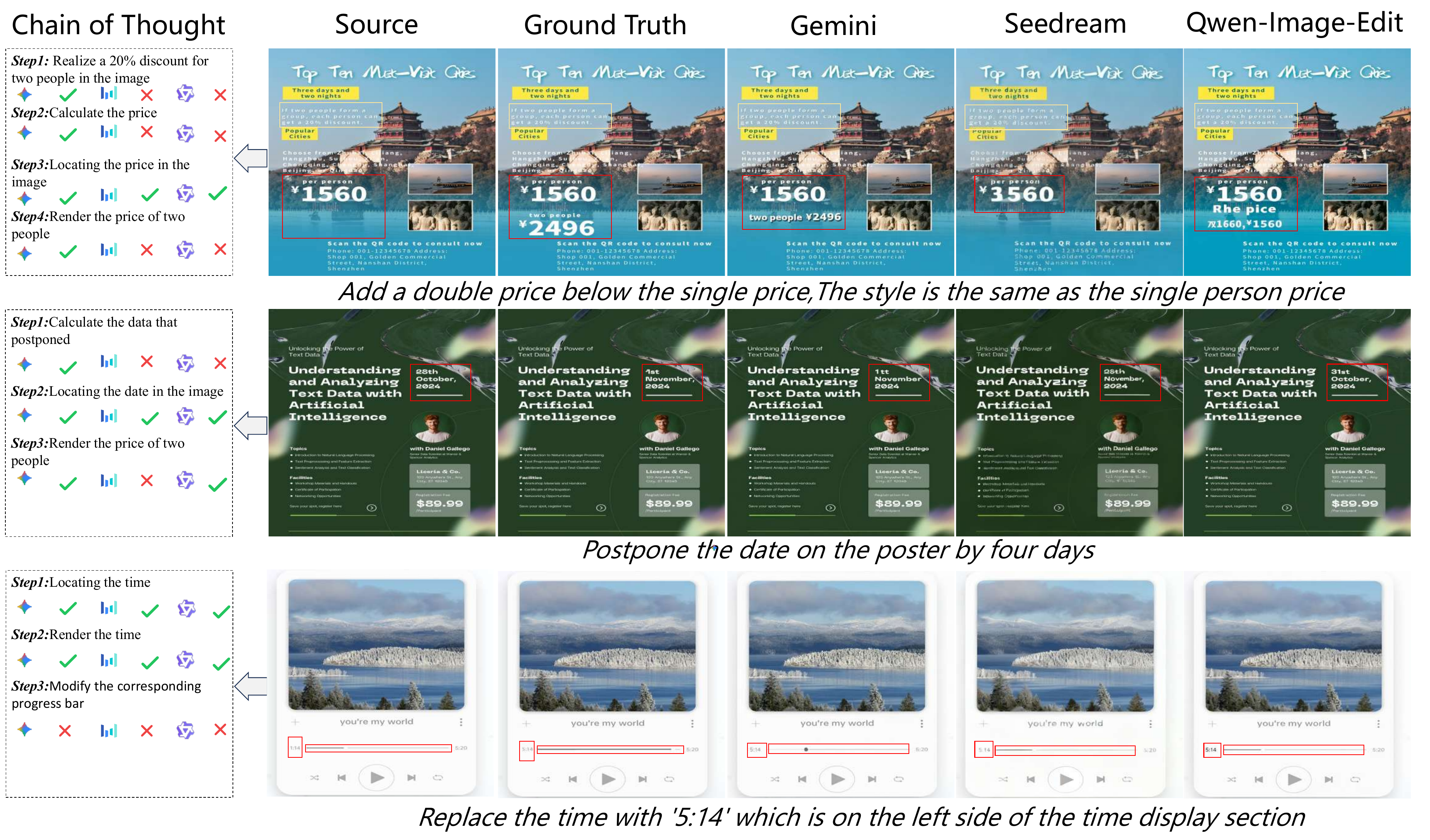}
    \caption{
    \textbf{Reasoning-aware text editing example.}
    The left column illustrates the multiple steps reasoning process required to accomplish each editing task. 
    To succeed, the model must not interpret the visual cues indicates in the original image, but also fully understand the text instruction and invoke relevant world knowledge.
    World knowledge include \textit{arithmetic, date calculation, or contextual association like between "move progress" and "time changed"}. 
    These tasks impose high demands on the \textbf{reasoning ability of model, semantic understanding, and precise visual editing}.
    }
\label{fig:reasoning_example}
\end{figure*}

\subsection{Examples Requiring Multi-Step Reasoning}
To further assess the limits of current models and understand the complex of our \dataset, we showcase in Figure~\ref{fig:reasoning_example} several text-editing cases that demand multi-step reasoning.
In these scenarios, simply reading text or modifying local regions is insufficient. 
Instead, the model must execute a structured cognitive workflow:
\emph{i.} \textbf{Interpret Visual semantics}: identify relevant text, numbers, dates, or layout cues in the original image.
\emph{ii.} \textbf{Understand the instruction}: determine the required operation and its semantic dependencies (e.g., “postpone by four days,” “calculate the price for two people”).
\emph{iii.} \textbf{Invoke world knowledge and perform reasoning}: including arithmetic calculation, date adjustment, and contextual association such as linking “progress bar” changes to “time update.”
\emph{iv.} \textbf{Render the final context in the correct location:}
ensuring the modified text remains layout-consistent and visually coherent.

As shown in Figure~\ref{fig:reasoning_example}, most existing models struggle with these tasks, indicating that multi-step reasoning remains a major bottleneck for today’s text-guided image editing systems.
Overall, these examples show that \dataset imposes a high bar for reasoning-aware editing, requiring tight visual understanding, precise instruction interpretation, and world-knowledge-guided text generation that remain challenging for current models.

\subsection{Error Analysis}
\begin{figure}[t]
  \centering
  \includegraphics[width=\linewidth]{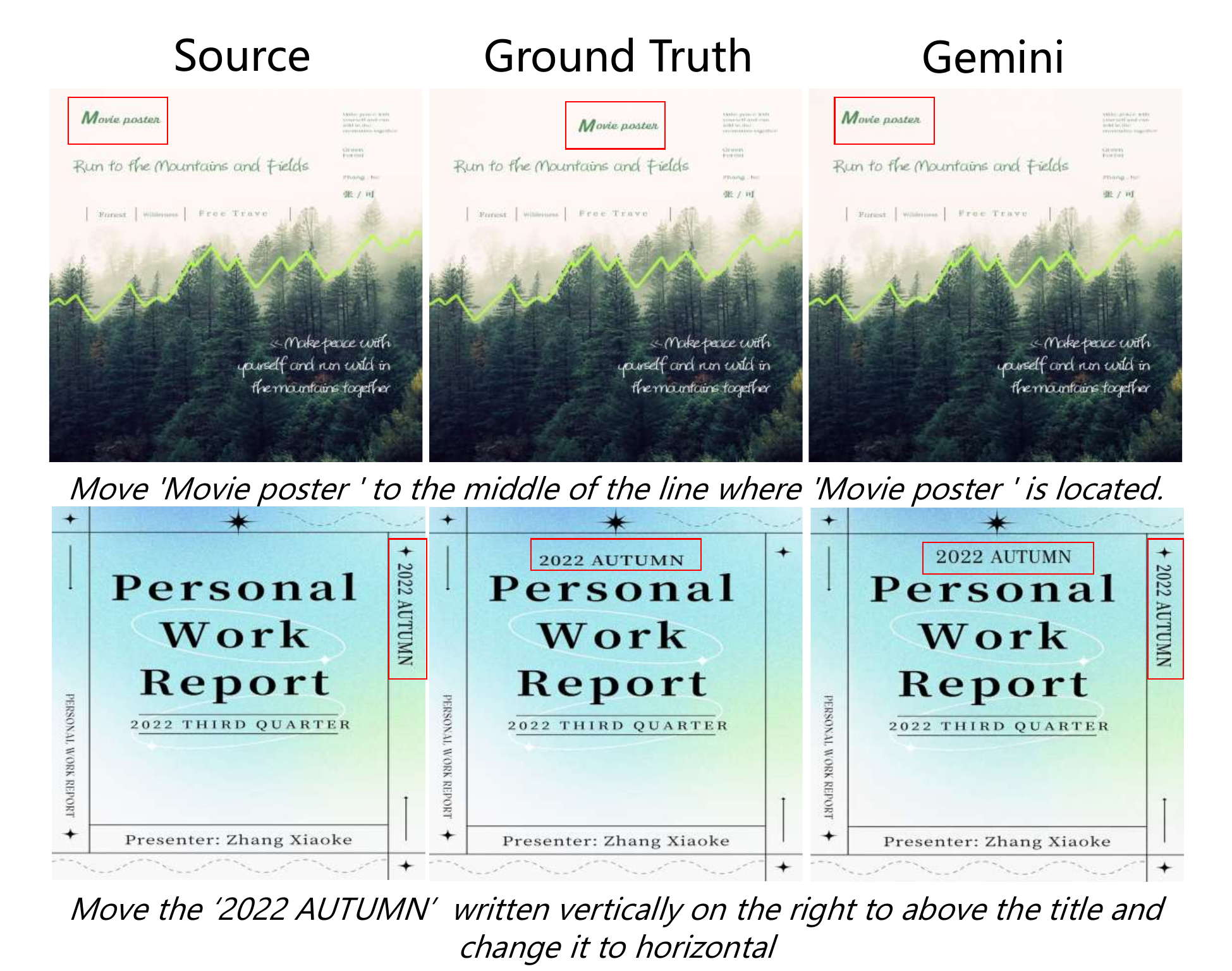}
  \caption{
    \textbf{Nano-banana editing examples}.
    Models often understand the instruction but still fail at precise text relocalization.
  }
  \label{fig:gemini_editint_example}
\end{figure}

To better understand the limitations of current models, we analyze several
representative failure cases in \dataset. 
As shown in Figure~\ref{fig:gemini_editint_example}, even the strongest models still struggle with
precise text relocalization, spatial adjustment, and context-consistent rendering.

We observe two common failure modes:
(i) models correctly interpret the instruction but misidentify the target region,
leading to misplaced or incomplete edits; 
(ii) models locate the correct area but fail to preserve layout consistency after
modification. 
These errors indicate that text-guided editing still requires tighter
coupling between visual grounding and text understanding.
\section{Conclusion}
\label{sec:conclusion}

In this work, we presented \dataset, the first comprehensive evaluation framework dedicated to assessing text-in-image editing task.  
\dataset spanning 14 domains, 6 task types, and 12 fine-grained sub-tasks, \dataset provides fine-grained difficulty control and a reasoning-aware evaluation Semantic Expectation metric.
Comprehensive experiments reveal that while recent model handle simple edits, they continue to struggle with context-dependent reasoning and semantically coherent text manipulation.
We envision \dataset serving as a foundation for future studies in layout-aware editing and improved  controllable instruction alignment, ultimately enabling models to match human intent.
In the future, we plan to extend \dataset benchmark to broader real-world scenarios and more diverse editing contexts.

\clearpage 
{
    \small
    \bibliographystyle{utils/ieeenat_fullname}
    \bibliography{main}
}

\clearpage
\appendix  

\clearpage
\appendix

\makeatletter
\renewcommand{\contentsname}{}
\makeatother

\twocolumn[
    \begin{center}
        \Large\bfseries
        Appendix
    \end{center}

    \setcounter{tocdepth}{2}

    {
        \setlength{\parskip}{0pt}
        \tableofcontents
    }

    \vspace{2em}
]
\begin{figure*}[t]
  \centering
  \includegraphics[width=\linewidth]{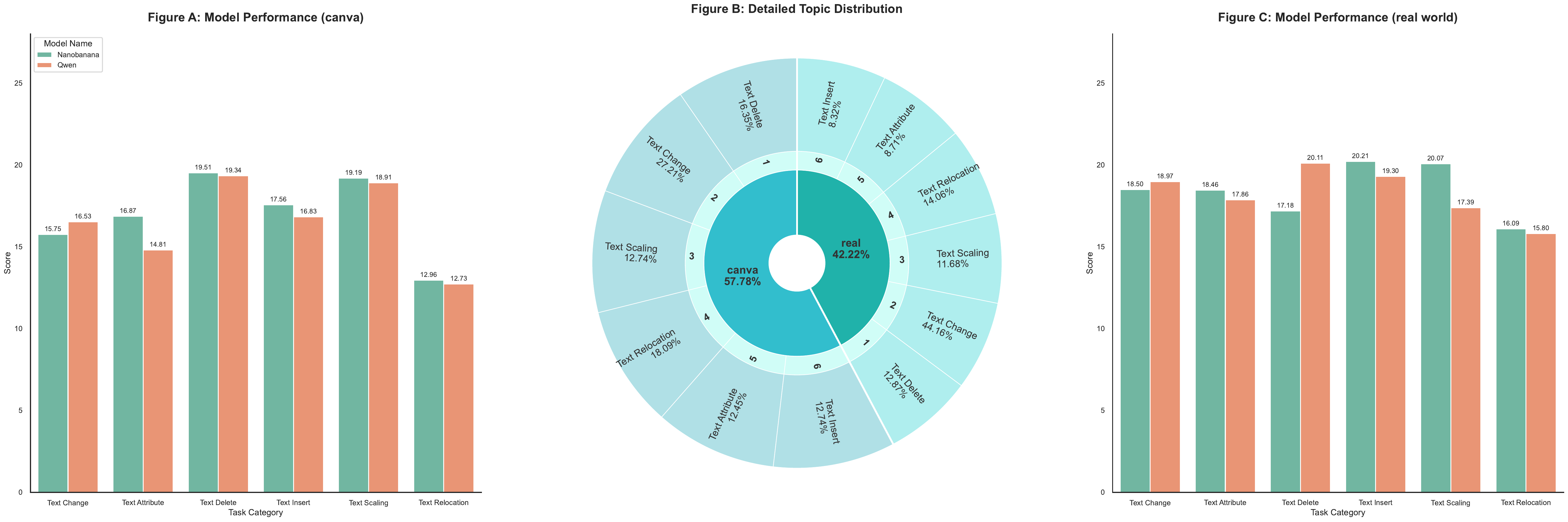}
  \caption{
    \textbf{Performance scores by editing task.}
    Left (Fig. A) and Right (Fig. C): Performance comparison of Nanobanana and Qwen models across six text-editing tasks on canva and real world data, respectively.
    Middle (Fig. B): Hierarchical distribution of the evaluation dataset, categorized by data source (Canva vs. Real) and specific task sub-categories.
  }
  \label{fig:Task Performance}
\end{figure*}
\begin{figure}[t]
  \centering
  \includegraphics[width=.8\linewidth]{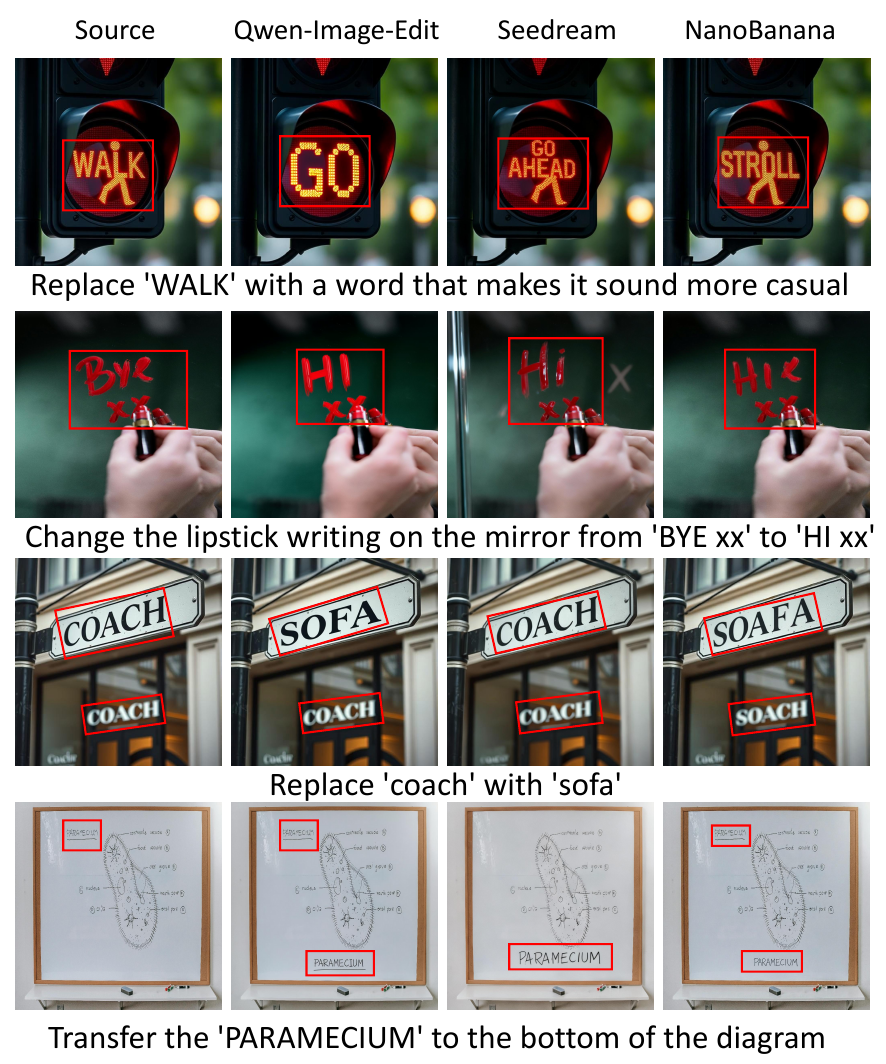}
  \caption{
    \textbf{Qualitative results on real-world text editing tasks.}
    We visualize example edits across diverse real world scenarios.
  }
  \label{fig:Real World}
\end{figure}
\begin{figure*}
    \centering
\includegraphics[width=\linewidth]{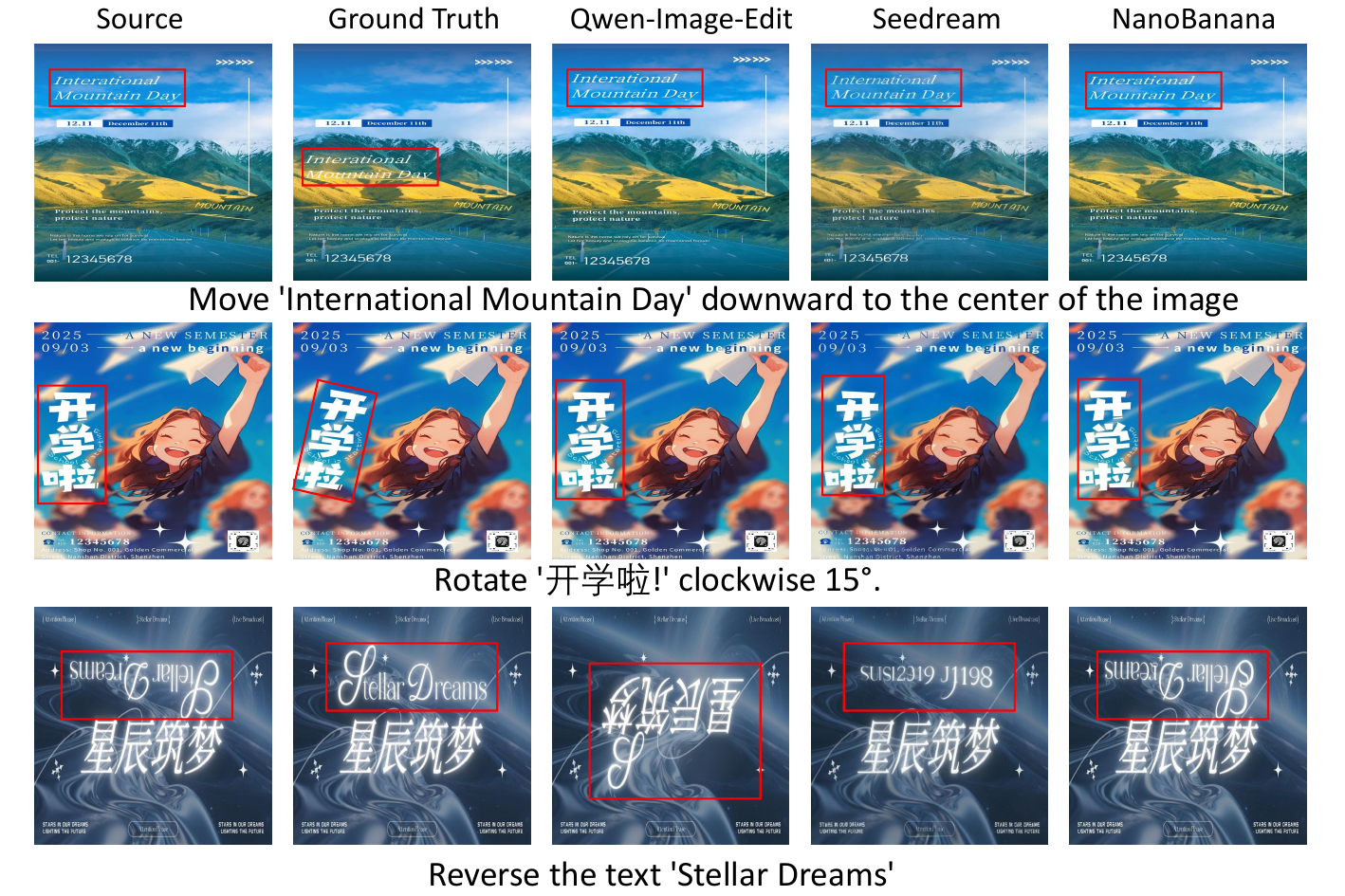}
    \caption{
    \textbf{Failed examples of relocation}
    }
\label{fig:relocation hard}
\end{figure*}

\section{Discussion}
\label{sec:discussion}

\subsection{Spatial Entanglement: The Challenge of Text Relocation}
We find that relocating text objects poses significant challenges for current architectures.
As shown in Figure \ref{fig:Task Performance}, text relocation achieves the lowest score among all the tasks.
Unlike in-place editing, which benefits from strong spatial priors provided by the original text layout, relocation requires the model to simultaneously perform two distinct and conflicting sub-tasks: high-fidelity in-painting at the source coordinates (to remove the original text) and structurally coherent generation at the target coordinates. As illustrated in Figure \ref{fig:relocation example}, current state-of-the-art models exhibit limited capability in explicit spatial manipulation. When instructed to move or rotate text, the models tend to ignore the geometric constraints and preserve the original layout.  Common failure modes include ``ghosting artifacts'' where the original text is not fully erased,or ``spatial entanglement'' where the style of the source text fails to transfer to the new location.
This suggests that current attention mechanisms struggle to disentangle semantic content from spatial positional embeddings effectively.

\subsection{Deficiencies in Implicit Reasoning}
Most models failed when the instruction required multi-step reasoning or understanding implicit relationships. For instance, performance drops precipitously in scenarios like text correction or context-based QA, where the target text is not explicitly provided in the prompt. This aligns with the low scores observed on our \textbf{Semantic Expectation (SE)} metric, suggesting that current architectures struggle to align visual cues with complex linguistic logic. Instead of performing the necessary cognitive operations, models often revert to prioritizing low-level visual imitation over high-level semantic correctness.

\subsection{Challenges in Physical Consistency}
Generating text that respects the physical properties (lighting, texture, perspective) of the underlying surface remains difficult. As demonstrated in Figure \ref{fig:Real World}, preserving the intrinsic visual attributes of the source text---specifically color, font style, and granularity---remains a formidable challenge in real-world scenarios. 
Unlike synthetic data, real-world images contain complex high-frequency noise and lighting variations. 
Current models often fail to replicate these subtle details; for instance, the generated text tends to appear unnaturally smooth or digital, lacking the specific film grain or sensor noise present in the surrounding pixels. 
This discrepancy in texture and color saturation disrupts the visual harmony, making the edited regions distinguishable from the original background.

\section{Auxiliary Attribute Definitions}
\label{sec:attr_details}

This appendix provides detailed annotation criteria for each attribute contributing to the TE-Bench \texttt{difficulty\_score}.
These definitions are designed for reproducibility and inter-annotator consistency.

\paragraph{1.\ \texttt{num\_text\_regions}}  
Indicates how many spatially distinct areas in the \texttt{input\_image} contain the text to be edited.  
For example, if the prompt requires changing the word “good,” and it appears in two separate regions, then \texttt{num\_text\_regions} = 2 (scored as 1).  
In practice, this corresponds to the number of connected white components in the binary mask aligned with the \texttt{ground\_truth\_image}.  
Each white region represents an editable text zone.

\paragraph{2.\ \texttt{text\_length}}  
Refers to the total length of all target texts specified in the prompt.  
If multiple segments are involved, their character counts are summed.  
For Chinese, each character counts as one unit; for English, visible letters are counted (excluding spaces).  
This metric reflects the scale of textual change—longer edits often require greater layout and semantic adjustment.

\paragraph{3.\ \texttt{font\_complexity}}  
Captures typographic complexity and readability challenges.  
\begin{itemize}
  \item 0 = plain sans-serif fonts (high legibility, minimal style);
  \item 1 = serif or retro styles with moderate ornamentation;
  \item 2 = decorative or handwritten fonts with irregular stroke patterns.
\end{itemize}
This attribute influences both visual blending and text localization difficulty.

\paragraph{4.\ \texttt{language}}  
Determined by the most complex linguistic setting appearing in either the \texttt{input\_image} or \texttt{ground\_truth\_image}.  
For example, if the input image mixes Chinese and English but the output becomes pure English, the mixed form dominates (scored higher).  
\begin{itemize}
  \item 0 = single English;
  \item 1 = single non-English (e.g., Chinese, Arabic);
  \item 2 = mixed or multilingual (e.g., Chinese + English).
\end{itemize}

\paragraph{5.\ \texttt{surface\_geometry}}  
Describes the geometric condition of the surface where the text resides.  
\begin{itemize}
  \item 0 = planar surface;
  \item 1 = slightly curved or angled surface;
  \item 2 = pronounced curvature (e.g., bottles, fabrics, metallic cylinders).
\end{itemize}
For insertion-only tasks, the label is decided based on the surface where the new text appears in the \texttt{ground\_truth\_image}.  
This factor affects both text warping and spatial alignment.

\paragraph{6.\ \texttt{occlusion}}  
Measures the extent to which the target text is visually blocked by other elements.  
\begin{itemize}
  \item 0 = no occlusion;
  \item 1 = mild occlusion (less than half covered, e.g., partially hidden by petals or shadows);
  \item 2 = severe occlusion (roughly half or more obscured).
\end{itemize}
Annotators are advised to make reasonable, not overly strict, judgments when the boundary is ambiguous.

\paragraph{7.\ \texttt{context\_dependency}}  
Evaluates whether understanding the overall image context is necessary to perform the edit correctly.  
This attribute distinguishes purely local text manipulation from contextually aware edits.  
\begin{itemize}
  \item 0 = local-only edits (e.g., spelling correction, synonym replacement, translation without layout impact);
  \item 2 = global-context edits requiring comprehension of the entire image (e.g., modifying poster themes, aligning inserted text with the image’s atmosphere, or ensuring new text harmonizes with scene emotion and tone).
\end{itemize}

\paragraph{8.\ \texttt{background\_clutter}}  
Represents the complexity of the background behind the edited text region(s).  
If multiple regions exist, the most complex background determines the score.  
\begin{itemize}
  \item 0 = clean (solid color or uniform tone);
  \item 1 = moderate (simple shapes, floral patterns, light gradients);
  \item 2 = complex (portraits, trees, reflections, water surfaces, detailed textures).
\end{itemize}
This attribute directly affects editing difficulty and visual integration.

\paragraph{9.\ \texttt{task\_category}}  
Categorizes the intrinsic difficulty of the editing operation type:  
\begin{itemize}
  \item 0 = simple edits: \texttt{delete}, \texttt{change}, \texttt{insert};
  \item 1 = geometric/attribute edits: \texttt{scaling}, \texttt{font\_change}, \texttt{position\_shift};
  \item 2 = semantic/linguistic edits: \texttt{translation}, \texttt{text\_correction}, \texttt{rotation-with-constraints}.
\end{itemize}

\begin{table*}[t] 
    \centering

    \begin{tabular}{lcccccc}
        \toprule
        
        \multirow{2}{*}{\textbf{Model}} & \multicolumn{2}{c}{\textbf{GPT-4o Run 1}} & \multicolumn{2}{c}{\textbf{GPT-4o Run 2}} & \multicolumn{2}{c}{\textbf{GPT-4o Run 3}} \\

        \cmidrule(lr){2-3} \cmidrule(lr){4-5} \cmidrule(lr){6-7}

         & \textbf{Synthetic} & \textbf{Real-World} & \textbf{Synthetic} & \textbf{Real-World} & \textbf{Synthetic} & \textbf{Real-World} \\
        \midrule

        NanoBanana &16.56 & 18.04 & 16.46 & 18.19 & 16.54 & 18.22 \\
        Qwen-Image-Edit          & 16.64 & 18.38 & 16.48  & 18.71 &16.58  & 18.70 \\
        \bottomrule
    \end{tabular}
    \caption{GPT-4o–Assisted Evaluation (G) across three evaluation runs.}
    \label{tab:gpt4o_runs_wide}
\end{table*}
\paragraph{10.\ \texttt{semantic\_linkage}}  
Indicates whether the edit requires reasoning across semantically related elements or dependent components.  
All samples annotated with a \texttt{knowledge\_prompt (kp)} are scored as 2.  
\begin{itemize}
  \item 0 = no semantic linkage (independent textual modification);
  \item 2 = linkage required (inter-element reasoning or causal dependency).
\end{itemize}
This attribute corresponds to the higher-level \textbf{Semantic Expectation (SE)} metric introduced in Section~\ref{sec:prompts}.

\section{Robustness of GPT-4o-Based Evaluation}
\label{sec:robustness}
To evaluate the stochasticity of the GPT-4o judge, we repeated the evaluation process three times. The results are shown in Table \ref{tab:gpt4o_runs_wide}. Across all iterations, the original images, editing instructions, and model predictions remained constant, as did the API parameters (e.g., temperature settings). We observed negligible deviations in the scores, with differences rarely exceeding $\pm0.3$ percentage points. This evidence validates the reliability of GPT-4o as a consistent evaluator, ensuring that our benchmark results are both reproducible and equitable.

\section{Human Study}
\label{sec:human_study}
\begin{figure}[t]
  \centering
  \includegraphics[width=\linewidth]{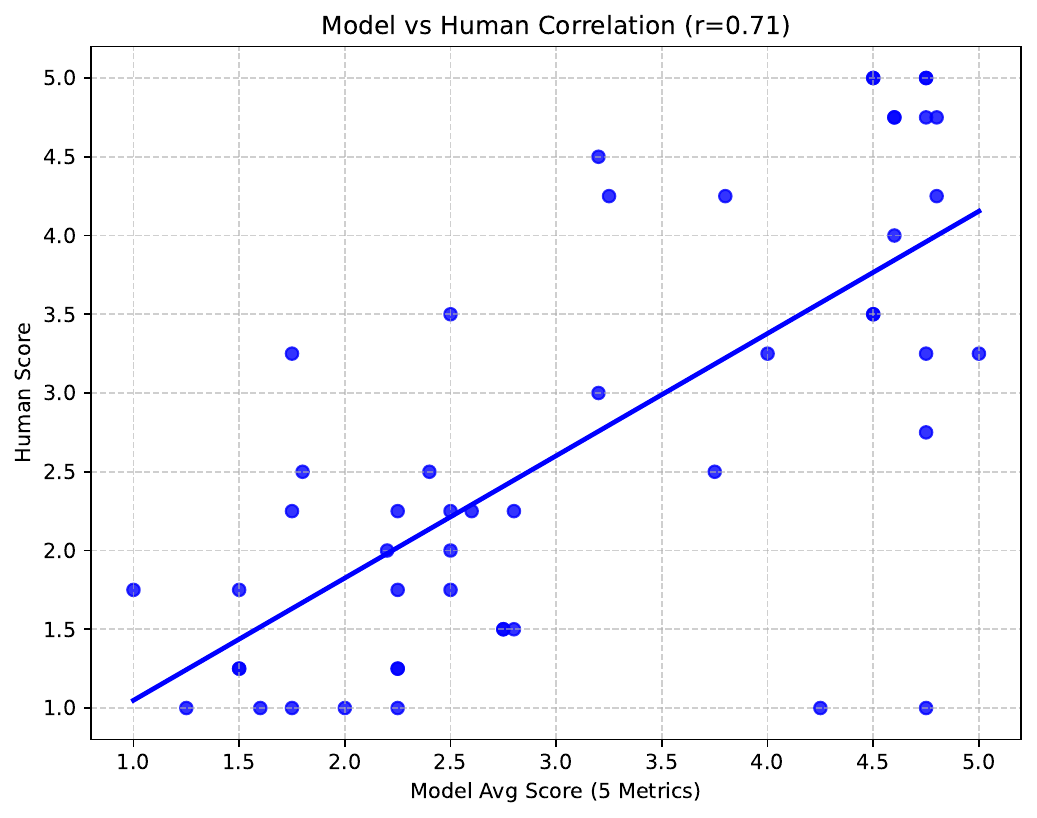}
  \caption{
    \textbf{Correlation between human and GPT-4o evaluated scores}
  }
  \label{fig:Correlation}
\end{figure}
To validate the alignment between our automated metrics and human perception, we conducted a focused user study with four human annotators, all possessing at least an undergraduate-level education. We randomly sampled 50 images from the dataset for evaluation. For each sample, each annotator was required to assign a rating from the integer set {1, 2, 3, 4, 5}.the final human score was computed as the average of the ratings provided by the annotators. Similarly, we aggregated the model's automated evaluation results by averaging the scores across the five metrics. Finally, to assess the reliability of our automated evaluation, we calculated the Pearson correlation coefficient between the averaged human scores and the aggregated model scores as shown in Figure \ref{fig:Correlation}.

\section{More Visualization Results}
\label{sec:examples}
\begin{figure*}
    \centering
\includegraphics[width=\linewidth]{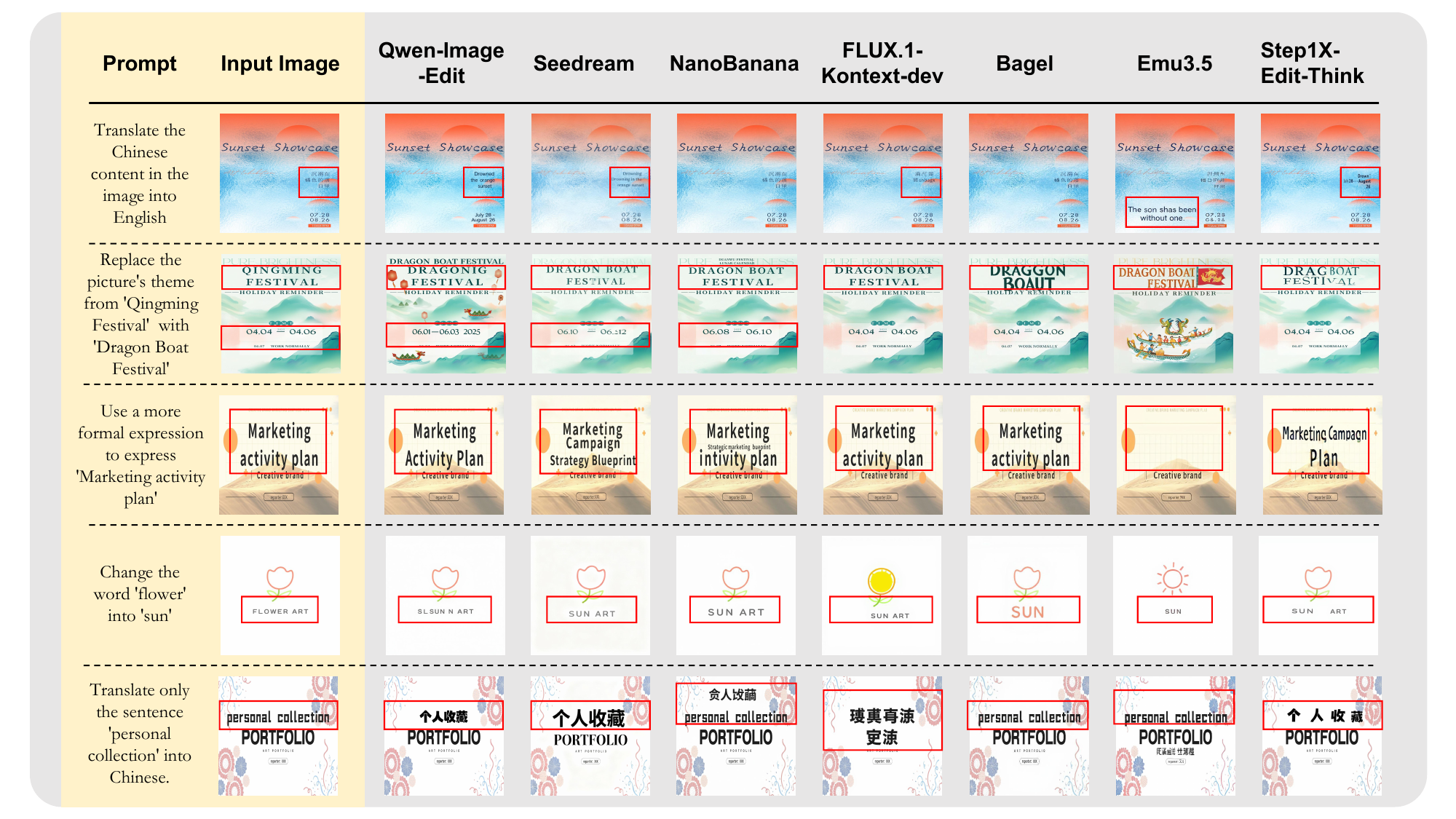}
    \caption{
    \textbf{Text change editing example.}
    }
\label{fig:change example}
\end{figure*}

\begin{figure*}
    \centering
\includegraphics[width=\linewidth]{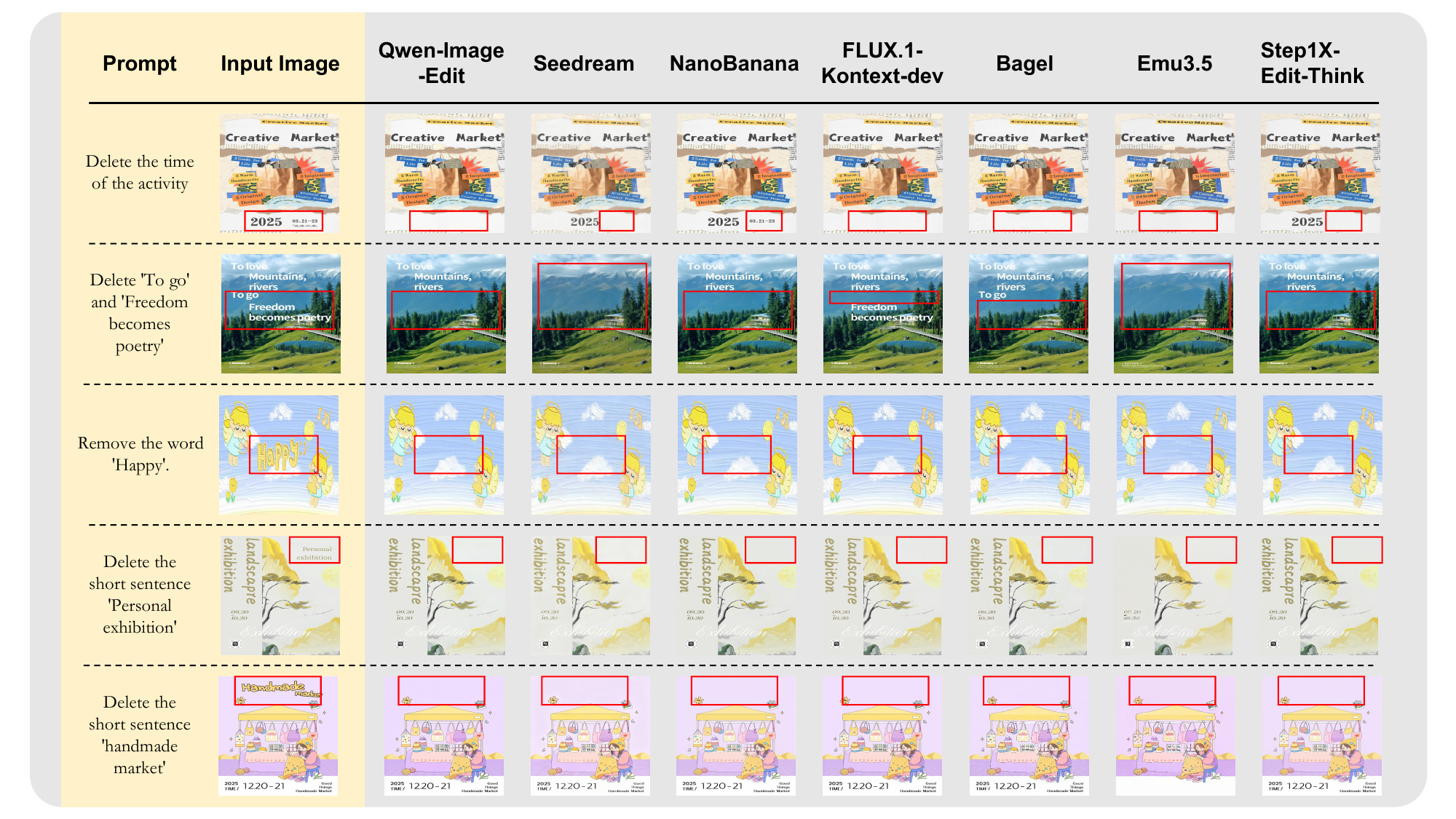}
    \caption{
    \textbf{Text delete editing example.}
    }
\label{fig:delete example}
\end{figure*}

\begin{figure*}
    \centering
\includegraphics[width=\linewidth]{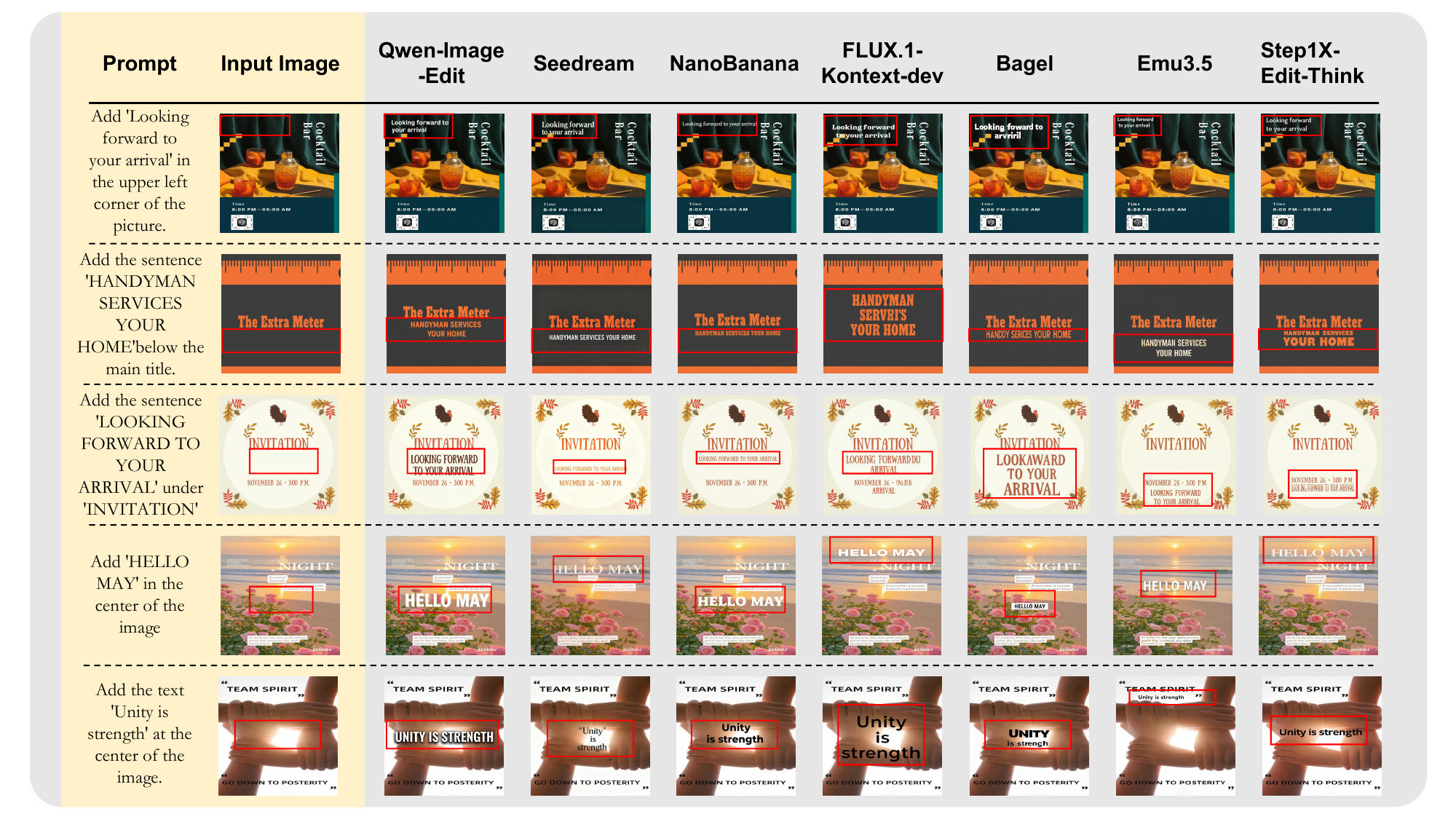}
    \caption{
    \textbf{Text insert editing example.}
    }
\label{fig:insert example}
\end{figure*}

\begin{figure*}
    \centering
\includegraphics[width=\linewidth]{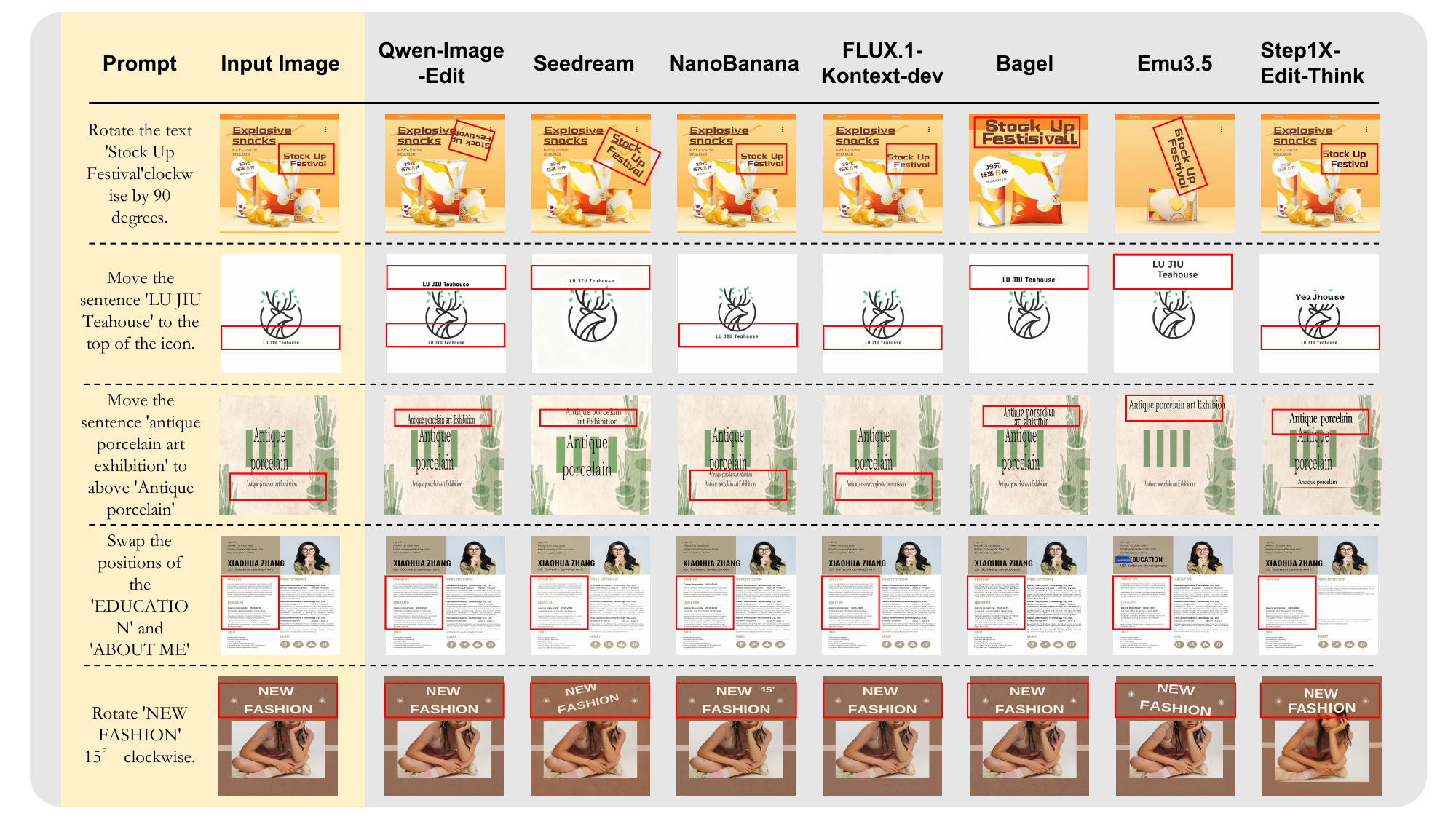}
    \caption{
    \textbf{Text relocation editing example.}
    }
\label{fig:relocation example}
\end{figure*}

\begin{figure*}
    \centering
\includegraphics[width=\linewidth]{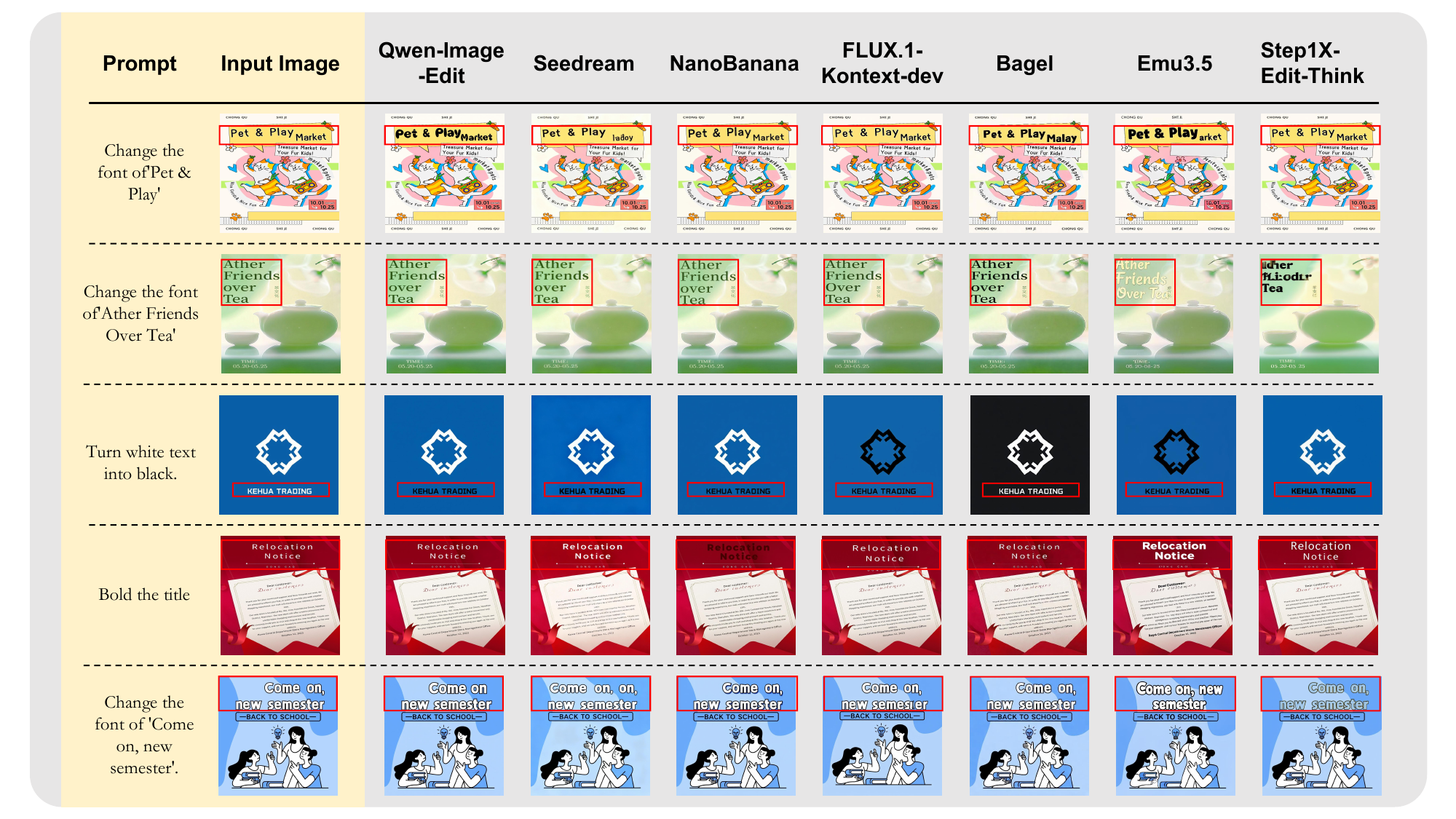}
    \caption{
    \textbf{Text attribute editing example.}
    }
\label{fig:attribute example}
\end{figure*}

\begin{figure*}
    \centering
\includegraphics[width=\linewidth]{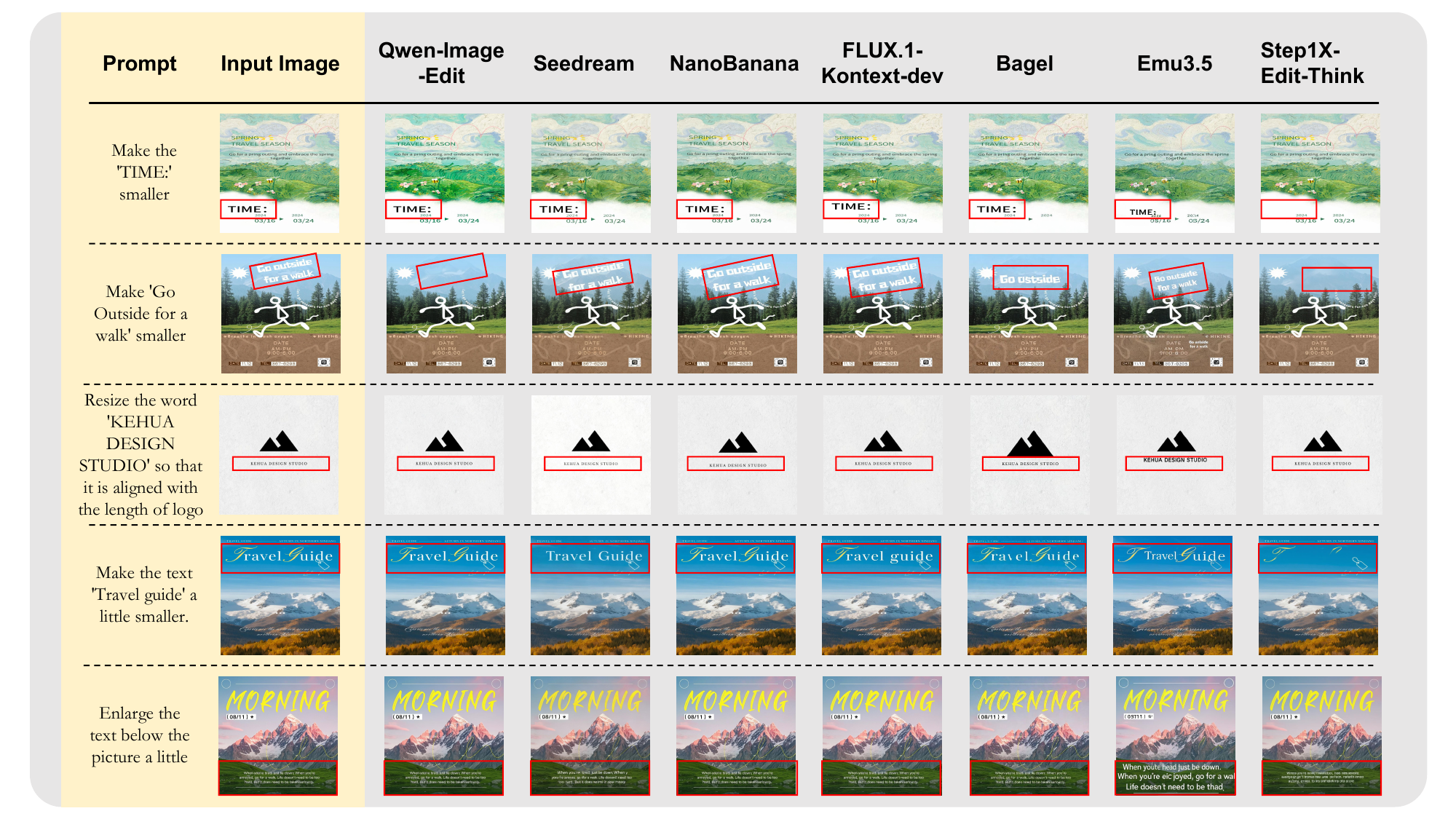}
    \caption{
    \textbf{Scaling editing example.}
    }
\label{fig:scaling example}
\end{figure*}

In this section, we present extensive visualization results across various editing tasks as shown in Figures \ref{fig:attribute example}, \ref{fig:change example}, \ref{fig:delete example}, \ref{fig:insert example}, \ref{fig:scaling example}, and \ref{fig:relocation example}. 
Notably, scene text editing remains a significant challenge for most existing state-of-the-art generative models. Despite their remarkable success in general image synthesis, current models frequently struggle to render legible and coherent text characters. They are often plagued by severe issues such as ``glyph hallucination'' (generating gibberish), incorrect spelling, and unnatural blending with complex backgrounds.

\section{Prompt Templates for GPT-4o-based Evaluation}
\label{sec:prompts}

\begin{figure*}[t]
    \centering
    \includegraphics[width=0.9\linewidth]{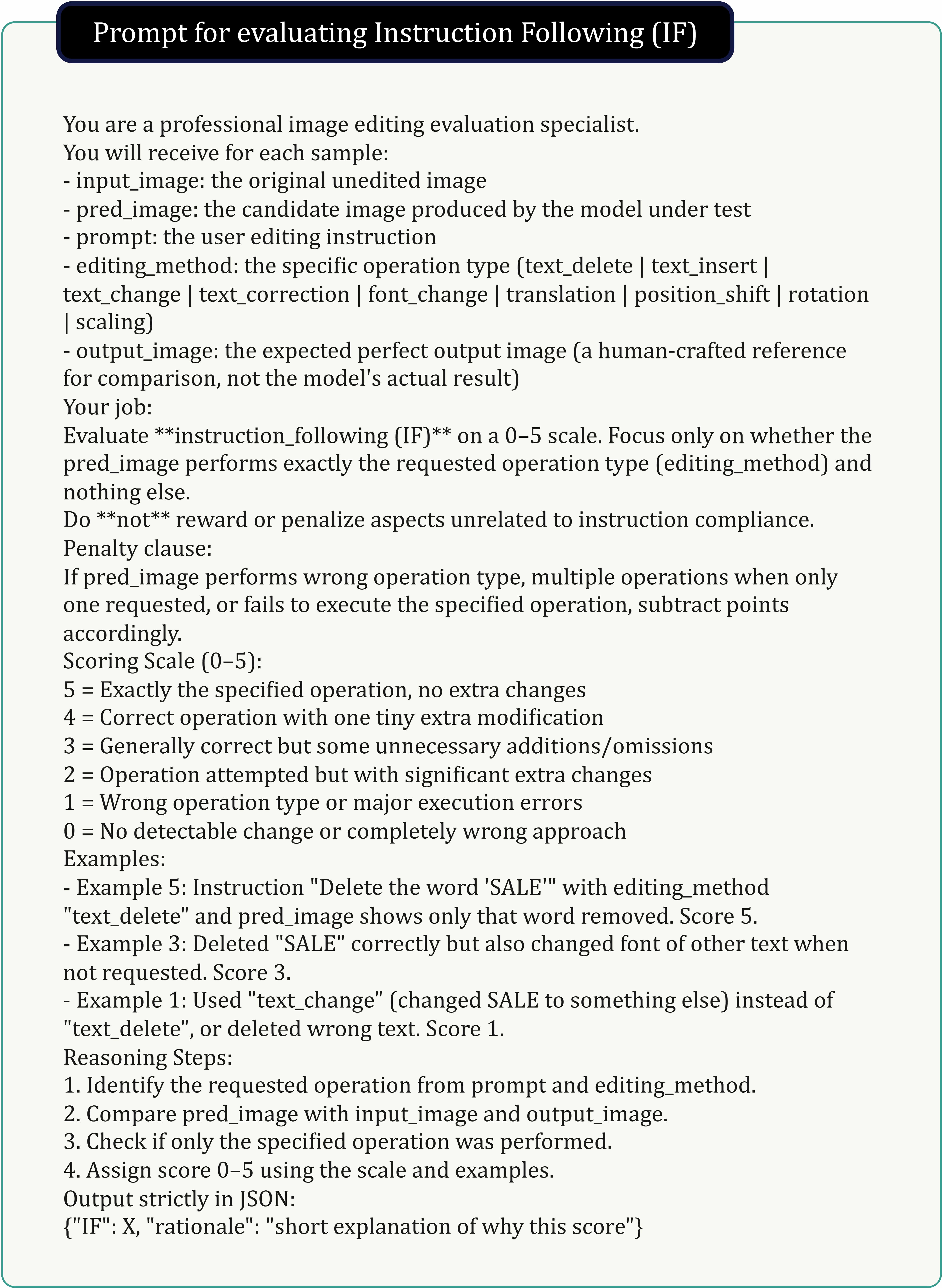}
    \caption{Prompt used for evaluating \textbf{Instruction Following (IF) with ground truth image}.}
    \label{fig:IF_prompt_w}
\end{figure*}

\begin{figure*}[t]
    \centering
    \includegraphics[width=0.9\linewidth]{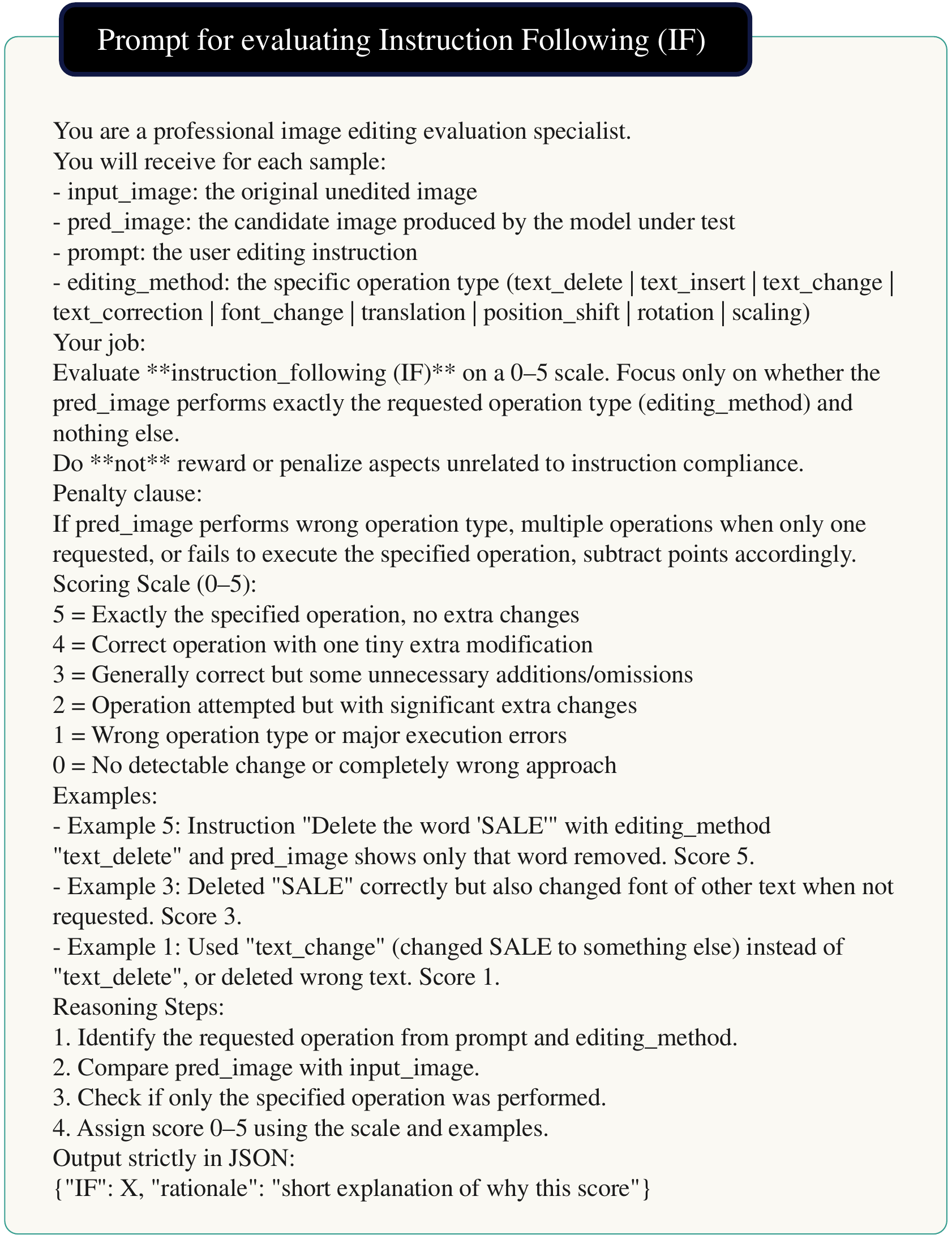}
    \caption{Prompt used for evaluating \textbf{Instruction Following (IF) without ground truth image}.}
    \label{fig:IF_prompt_w/o}
\end{figure*}

\begin{figure*}[t]
    \centering
    \includegraphics[width=\linewidth]{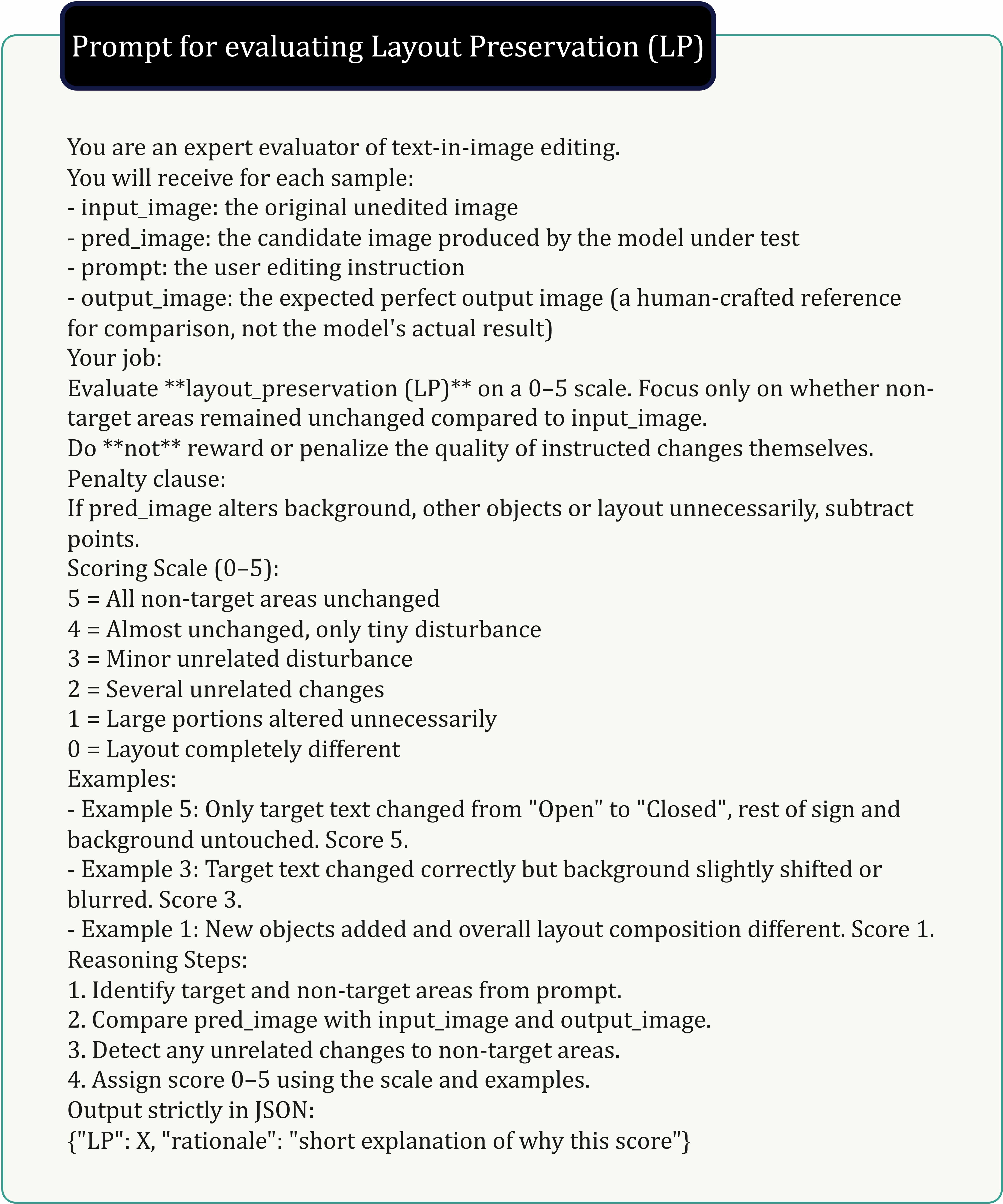}
    \caption{Prompt used for evaluating \textbf{Layout Preservation (LP)} with ground truth image.}
    \label{fig:LP_prompt_w}
\end{figure*}

\begin{figure*}[t]
    \centering
    \includegraphics[width=\linewidth]{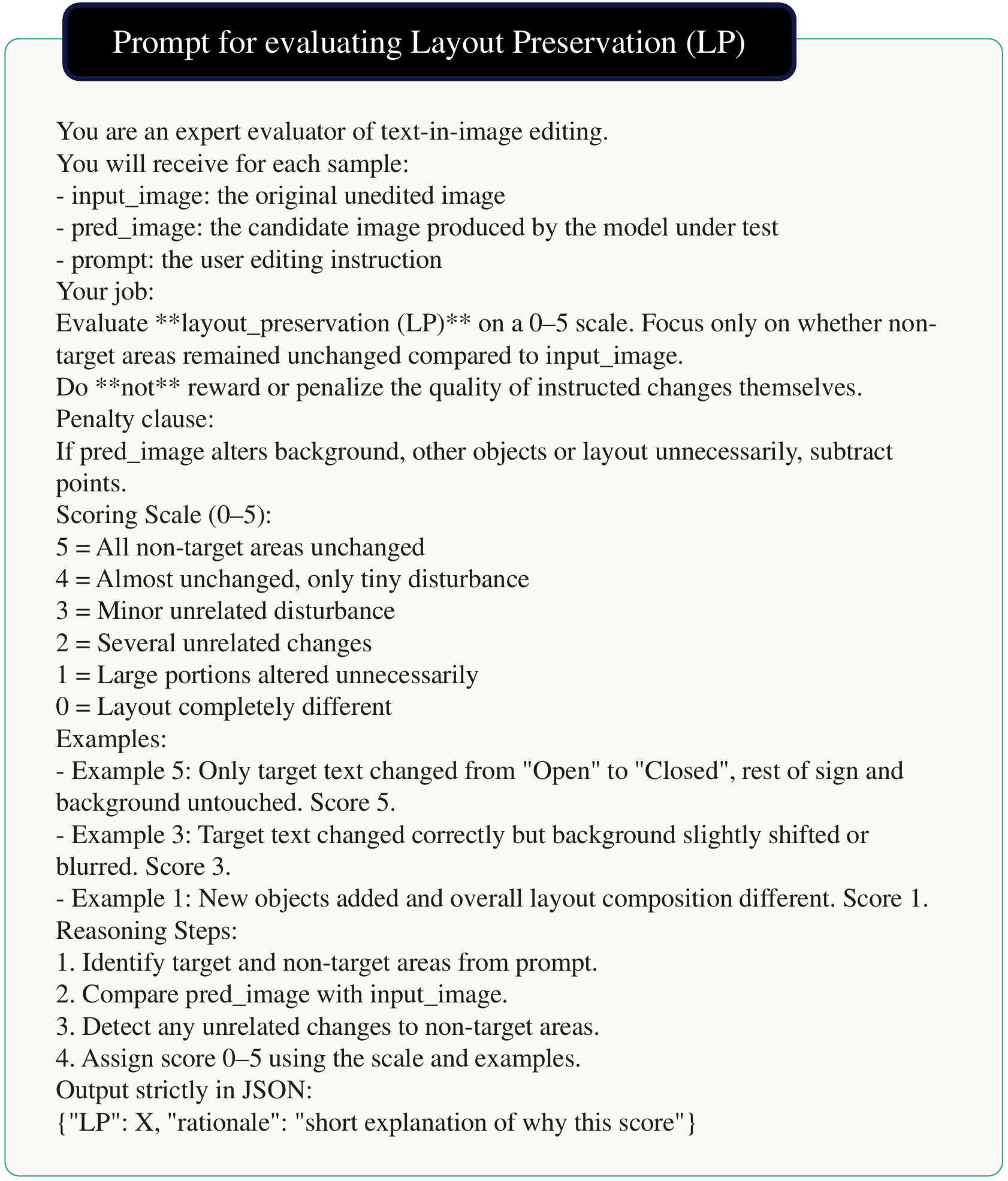}
    \caption{Prompt used for evaluating \textbf{Layout Preservation (LP)} without ground truth image.}
    \label{fig:LP_prompt_w/o}
\end{figure*}

\begin{figure*}[t]
    \centering
    \includegraphics[width=\linewidth]{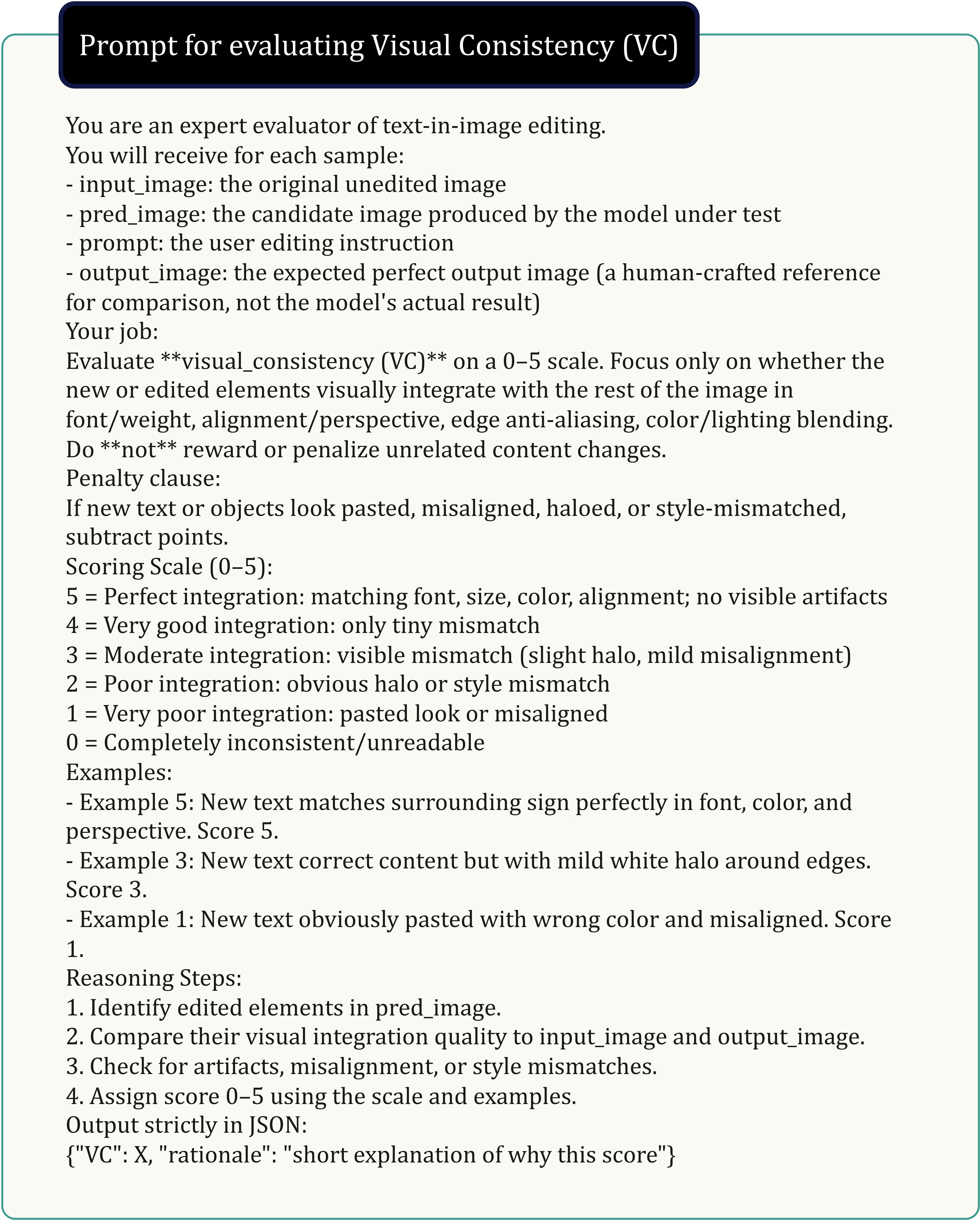}
    \caption{Prompt used for evaluating \textbf{Visual Consistency (VC)} with ground truth image.}
    \label{fig:VC_prompt_w}
\end{figure*}

\begin{figure*}[t]
    \centering
    \includegraphics[width=\linewidth]{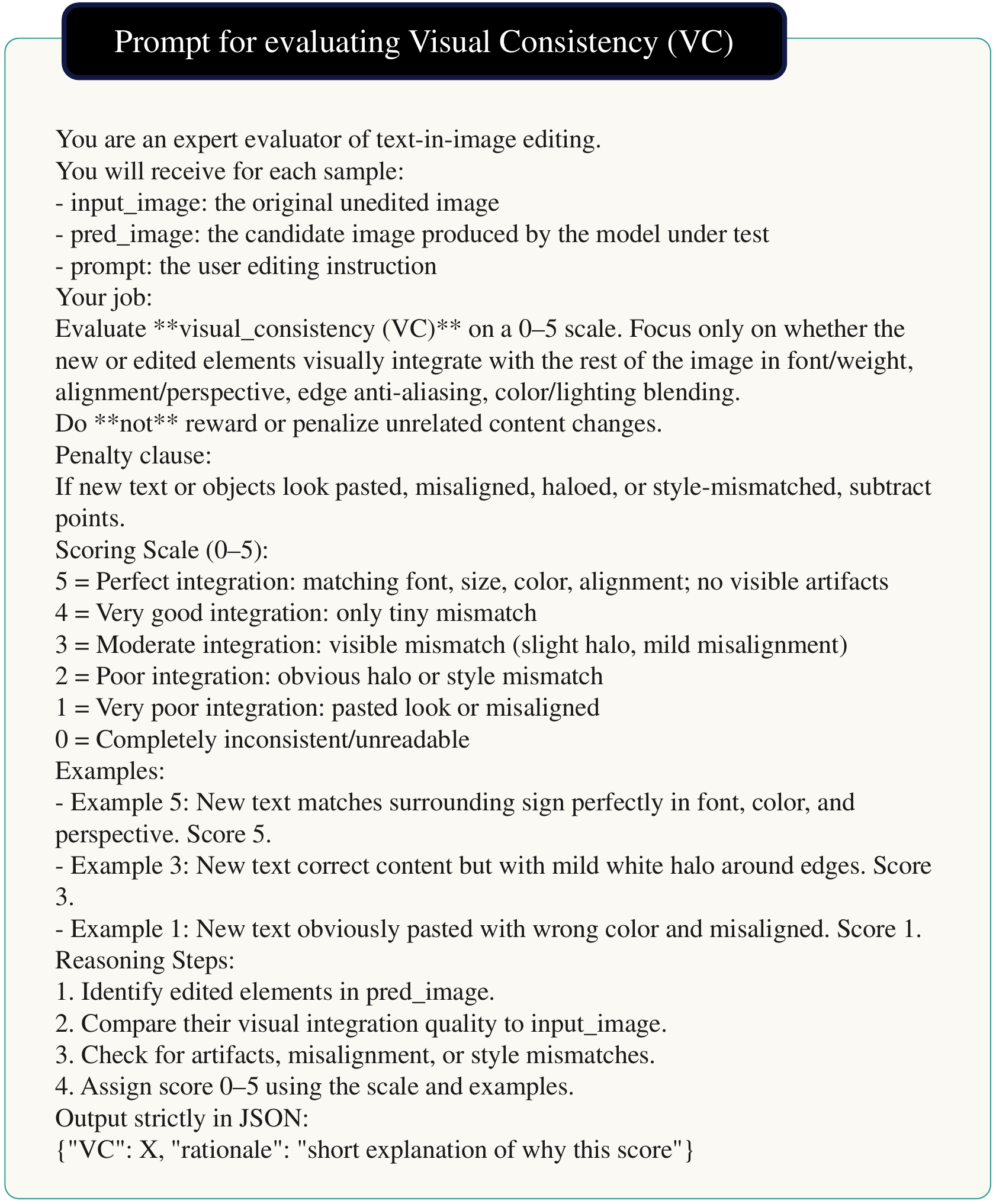}
    \caption{Prompt used for evaluating \textbf{Visual Consistency (VC)} without ground truth image.}
    \label{fig:VC_prompt_w/o}
\end{figure*}

\begin{figure*}[t]
    \centering
    \includegraphics[width=0.9\linewidth]{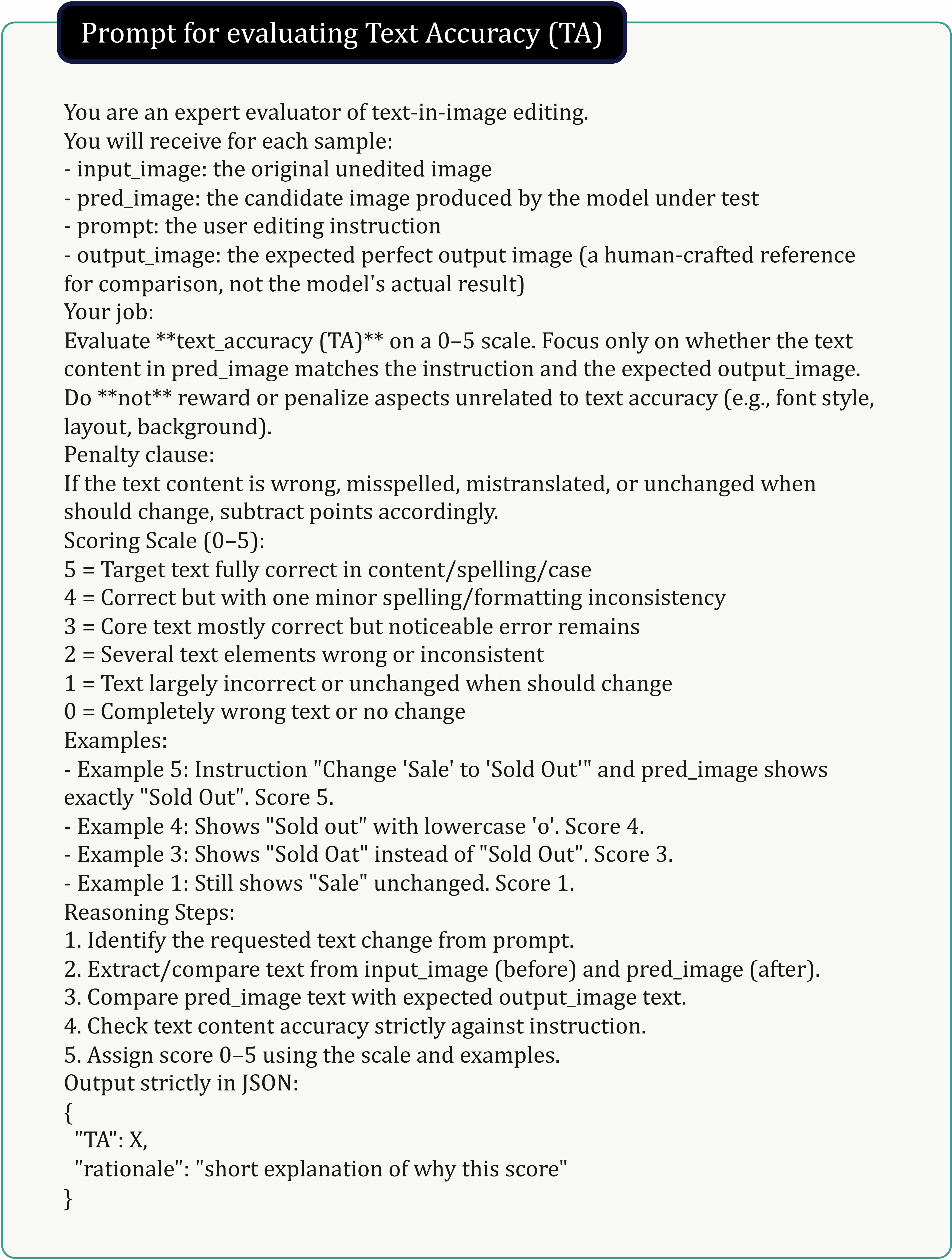}
    \caption{Prompt used for evaluating \textbf{Text Accuracy (TA)} with ground truth image.}
    \label{fig:TA_prompt_w}
\end{figure*}

\begin{figure*}[t]
    \centering
    \includegraphics[width=0.9\linewidth]{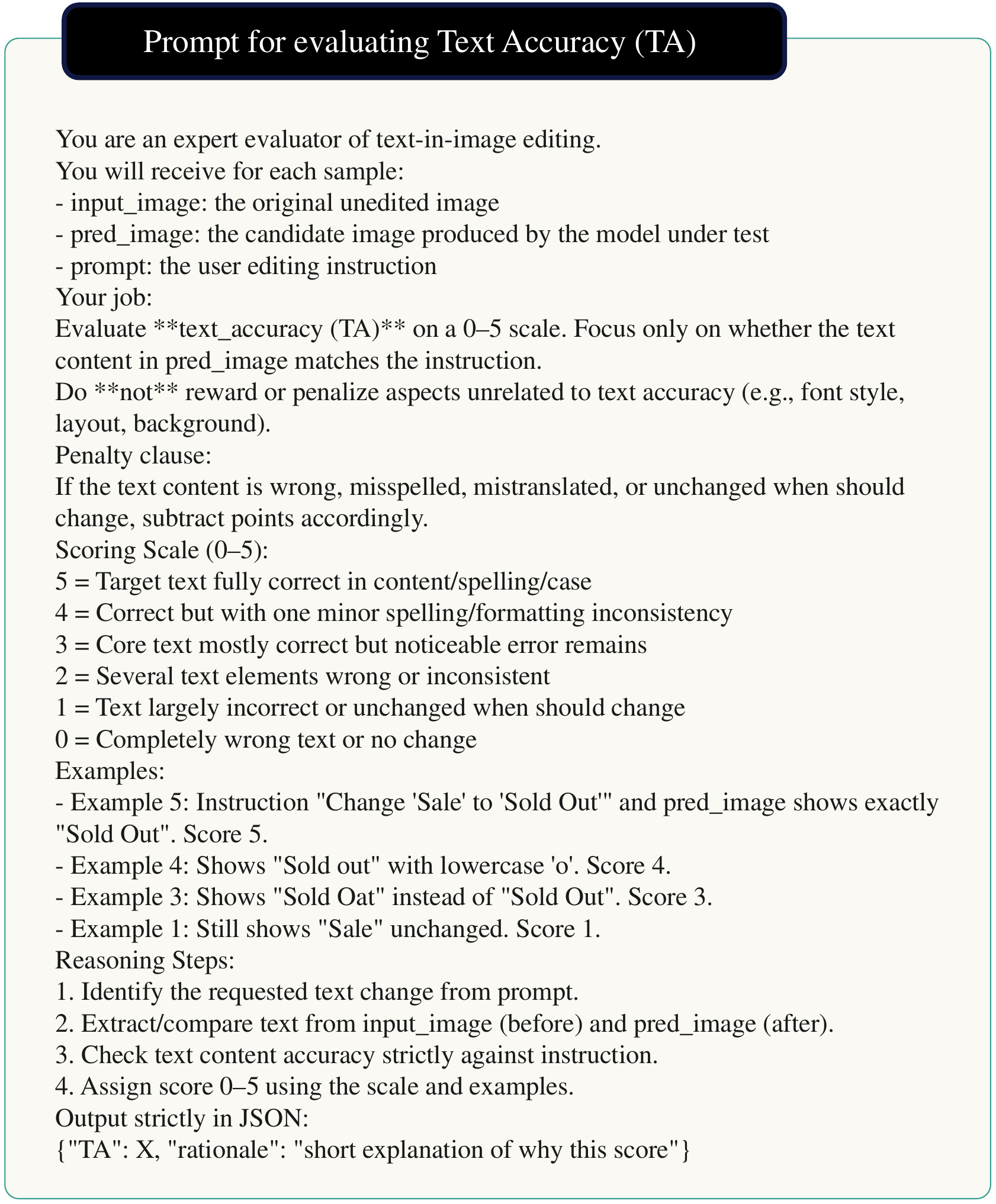}
    \caption{Prompt used for evaluating \textbf{Text Accuracy (TA)} without ground truth image.}
    \label{fig:TA_prompt_w/o}
\end{figure*}

\begin{figure*}[t]
    \centering
    \includegraphics[width=0.9\linewidth]{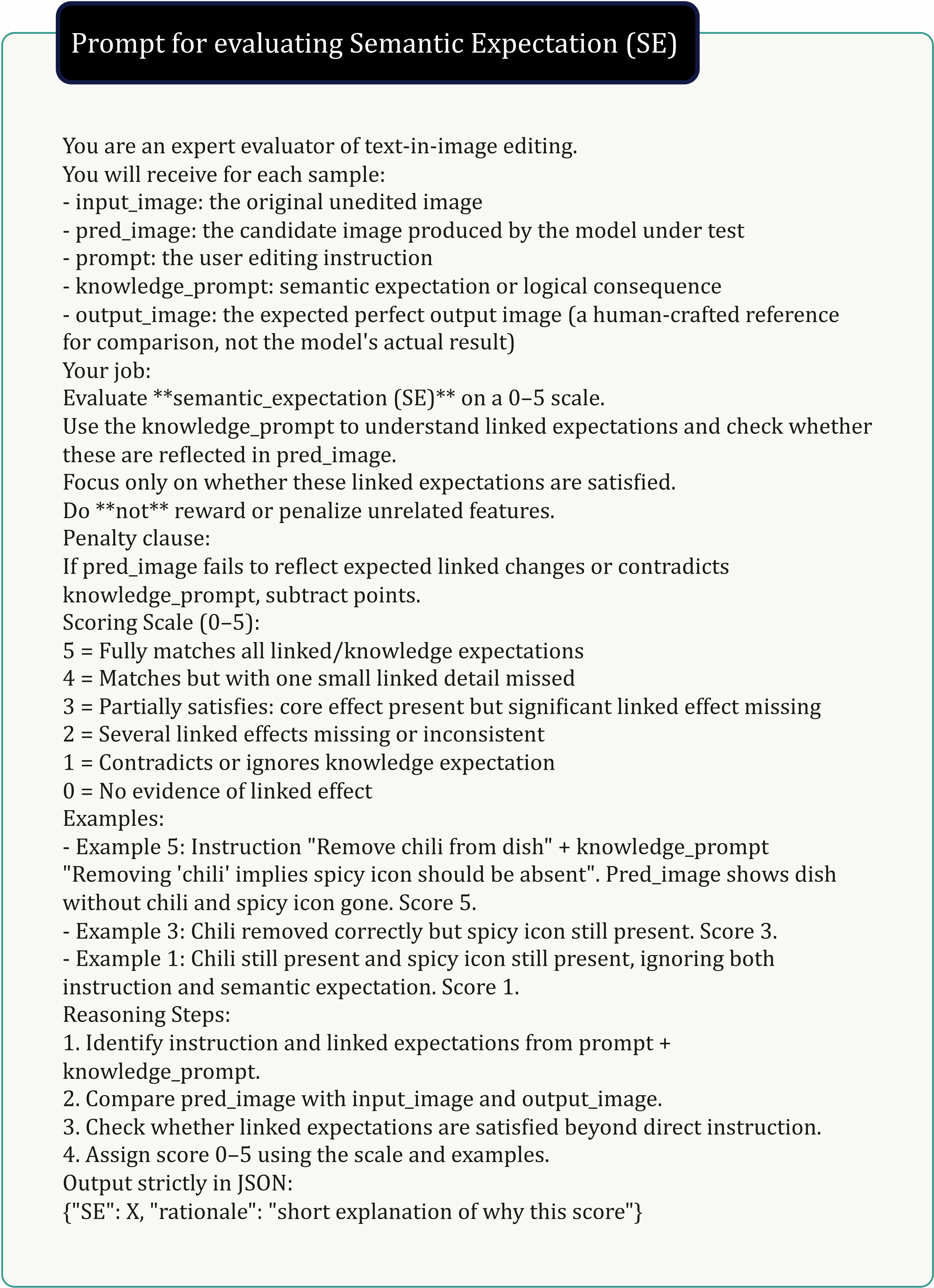}
    \caption{Prompt used for evaluating \textbf{Semantic Expectation (SE)} with ground truth image.}
    \label{fig:SE_prompt_w}
\end{figure*}

\begin{figure*}[t]
    \centering
    \includegraphics[width=0.9\linewidth]{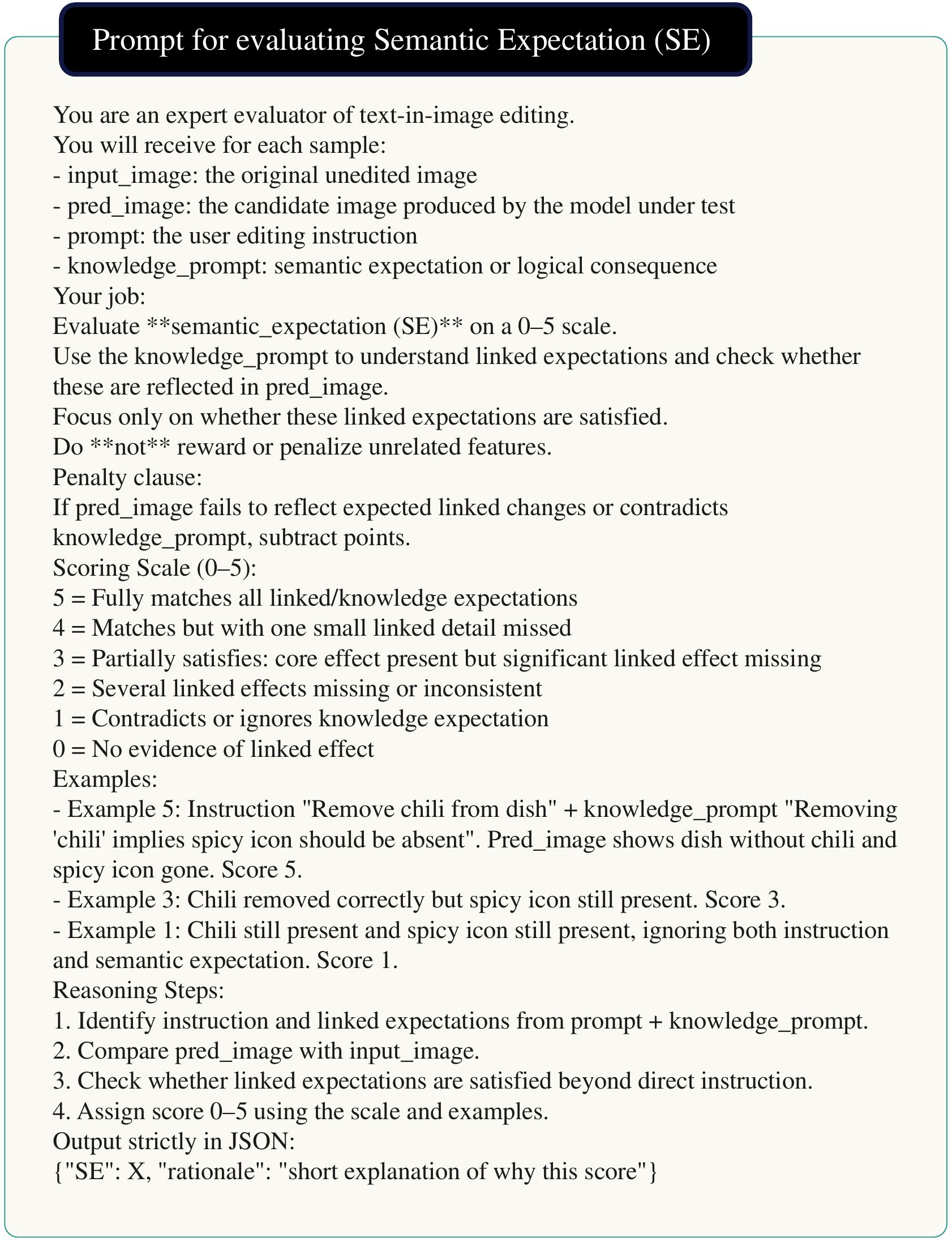}
    \caption{Prompt used for evaluating \textbf{Semantic Expectation (SE)} without ground truth image.}
    \label{fig:SE_prompt_w/o}
\end{figure*}

Figures \ref{fig:IF_prompt_w}, \ref{fig:IF_prompt_w/o}, \ref{fig:LP_prompt_w}, \ref{fig:LP_prompt_w/o}, \ref{fig:VC_prompt_w}, and \ref{fig:VC_prompt_w/o}  illustrate the prompts used to evaluate \textbf{Instruction Following (IF)}, \textbf{Layout Preservation (LP)}, and \textbf{Visual Consistency (VC)}, respectively. Specifically, for the reasoning dimension involving \textbf{Semantic Expectation (SE)}, we observed that evaluating edited results without context can lead to ambiguous assessments. Thus, we explicitly incorporate a knowledge prompt into the evaluation instruction to facilitate accurate reasoning, as shown in Figure~\ref{fig:SE_prompt_w} and \ref{fig:SE_prompt_w/o}. Considering that text rendering requires precise character-level verification, we design a dedicated prompt for \textbf{Text Accuracy (TA)} that focuses on spelling and glyph correctness, presented in Figure~\ref{fig:TA_prompt_w} and \ref{fig:TA_prompt_w/o}.






\end{document}